\documentclass[preprint]{agujournal}
\draftfalse

\usepackage{amsfonts,amssymb}
\usepackage{amsmath}
\usepackage{float}
\usepackage{bm}
\usepackage{url}
\usepackage{algorithm}
\usepackage[noend]{algpseudocode}
\usepackage{color}

\journalname{Water Resources Research}

\begin{document}

\title{Deep convolutional encoder-decoder networks for uncertainty quantification of dynamic multiphase flow in heterogeneous media}

\authors{Shaoxing Mo\affil{1,2}, Yinhao Zhu\affil{2}, Nicholas Zabaras\affil{2}, Xiaoqing Shi\affil{1}, and Jichun Wu\affil{1}}

\affiliation{1}{Key Laboratory of Surficial Geochemistry of Ministry of Education, School of Earth Sciences and Engineering, Nanjing University, Nanjing, China.}

\affiliation{2}{Center for Informatics and Computational Science, University of Notre Dame, Notre Dame, IN, USA.}

\correspondingauthor{Nicholas Zabaras}{nzabaras@gmail.com}
\correspondingauthor{Xiaoqing Shi}{shixq@nju.edu.cn}

\begin{keypoints}
\item A surrogate model for uncertainty quantification of dynamic multiphase flows is developed using deep convolutional encoder-decoder networks.
\item A two-stage network training strategy is proposed to improve the approximation of saturation discontinuity. 
\item The method can accurately predict the time-dependent output fields of a multiphase flow model with a 2500-D input permeability field.
\end{keypoints}

\begin{abstract}
Surrogate strategies are used widely for uncertainty quantification of groundwater models in order to  improve computational efficiency. However, their application to dynamic multiphase flow problems is hindered by the curse of dimensionality, the saturation discontinuity due to capillarity effects, and the time-dependence of the multi-output responses. In this paper, we propose a deep convolutional encoder-decoder neural network methodology to tackle these issues. The surrogate modeling task is transformed to an image-to-image regression strategy. This approach extracts high-level coarse features from the high-dimensional input permeability images using an encoder, and then refines the coarse features to provide the output pressure/saturation images through a decoder. A training strategy combining a regression loss and a segmentation loss is proposed in order to better approximate the discontinuous saturation field. To characterize the high-dimensional time-dependent outputs of the dynamic system, time is treated as an additional input to the network that is trained  using pairs of input realizations and of the corresponding system outputs at a limited number of time instances. The proposed method is evaluated using a geological carbon storage process-based multiphase flow model with a $2500$-dimensional stochastic permeability field. With a relatively small number of training data, the surrogate model is capable of accurately characterizing the spatio-temporal evolution of the pressure and discontinuous CO$_2$ saturation fields and can be used efficiently to compute the statistics of the system responses.

\end{abstract}

\section{Introduction}\label{intro}

Multiphase flow models are important tools in understanding the subsurface fluid flow processes involved in enhanced oil recovery, non-aqueous phase liquid pollution, and geological carbon storage (GCS). The model predictions (e.g., pressure and saturation) inherently involve some degree of uncertainty, which can result from the heterogeneity of the subsurface media property (e.g., permeability). The heterogeneous nature of the media, together with our incomplete knowledge about their properties, necessitate  modeling of these systems with stochastic partial differential equations (PDEs) with random input parameters~\citep{LIAO2017828,Luetal2016}. The random inputs to the stochastic PDEs hence lead to uncertain model predictions. Quantifying the influence of uncertainties associated with media properties on the model outputs is an indispensable task for facilitating science-informed decision making~\citep{Kitanidis2015,Luetal2016}. The Monte Carlo (MC) method is commonly used to address such   uncertainty quantification (UQ) problems~\citep{Ballio2004}. However,  the large number of realizations (i.e., simulation runs)   needed to obtain accurate statistics makes such approaches computationally impractical for computationally demanding multiphase flow models. 

Computationally efficient surrogate methods have achieved an increased popularity over the past decade~\citep{Asher2015,Razavi2012}. A surrogate model that is trained with a small number of model runs can provide an accurate and efficient approximation to the model input-output relationship. It can then be used as a full-replacement of the actual model when performing UQ tasks. Many surrogate methods based on the polynomial chaos expansion~\citep{xiu2002}, Gaussian processes~\citep{rasmussen2006}, and neural networks~\citep{HORNIK1989} have been applied widely to address UQ tasks in groundwater models with random inputs and have shown an impressive approximation accuracy and computational efficiency in comparison to MC  methods~\citep{CHAN2018,CREVILLENGARCIA20171,LI2009,LiaoZhang2013,LiaoZhang2014,LiaoZhang2016,LIAO2017828,MENG201713,MULLER20111527,TIAN2017113}.  

In this paper, we are concerned with transient multiphase flow problems in highly-heterogeneous subsurface media. For this kind of problems, three major challenges arise associated with the development of surrogate models. 

First, most existing surrogate methods fail to work when the input dimensionality increases~\citep{LIN2009712,MA20093084,MA20103884,LIAO2017828,TIAN2017113}. The high-dimensionality may result from the random heterogeneous media property (e.g. permeability), where often the input dimensionality is equal to the number of spatial grid points (pixels) used to define the property. One approach to alleviate the curse of dimensionality is to employ  dimensionality reduction techniques, such as the Karhunen-Lo{\` e}ve expansion~\citep{ZHANG2004773}, variational auto-encoders~\citep{LALOY2017387}, and generative adversarial networks~\citep{chan2017,LALOY2018}. These methods produce a low-dimensional latent representation of the random field. One then constructs a surrogate model from this latent space to the model outputs~\citep{Laloy2013,LI2009,LiaoZhang2013,LiaoZhang2014,LiaoZhang2016,LIAO2017828,LIN2009712,MENG201713,MULLER20111527}. However, since   realistic subsurface media properties are highly-heterogeneous, their reduced representation remains  high-dimensional ($> 100$). One can use adaptivity to further reduce the computational cost, in which the training samples for surrogate construction are adaptively selected, e.g.,  based on the importance of dimensions~\citep{GANAPATHYSUBRAMANIAN2007652,LIAO2017828} or local response nonlinearity~\citep{MA20093084,MO2017,ZHANG2013}. Such adaptive strategies can somewhat reduce the number of training samples, but the improvement is relatively limited for surrogate construction in problems of high-input dimensionality. 

Second, for multiphase flow problems, the saturation profile is discontinuous in the spatial domain due to  capillarity effects. This discontinuity leads to  one in the stochastic space as well making it extremely challenging to accurately approximate due to the fact that most surrogate models are continuous and often differentiable~\citep{Asher2015,MO2017,Xiu2005,LiaoZhang2013,LiaoZhang2014,LiaoZhang2016,LIAO2017828,LIN2009712}. A variety of approaches are proposed to handle discontinuities. For instance,~\citet{MA20093084,MA20103884} presented an adaptive hierarchical sparse grid method to locally refine discontinuous regions under the guidance of the hierarchical surplus.~\citet{LiaoZhang2013,LiaoZhang2014,LiaoZhang2016} proposed a series of transformed probabilistic collocation methods which approximate some alternative to the saturation variables having smooth relationship with the inputs.~\citet{LIAO2017828} proposed a two-stage adaptive stochastic collocation method in which the pressure and velocity were first generated by surrogate models and then substituted into the PDE to solve for the saturation. These methods even when they capture well the discontinuity remain prompt to the aforementioned curse of dimensionality.

Third, for dynamic problems,  the surrogate model should be able to predict efficiently the model outputs at arbitrary time instances. Constructing such a surrogate model is challenging. One common solution is to approximate the outputs at only one or several specific time instances~\citep [e.g.,][]{LI2009,LIN2009712,LiaoZhang2013,LiaoZhang2014,LiaoZhang2016,LIAO2017828}. Such methods will not be appropriate in capturing the time-dependence of the model response fields. 

In summary, the three aforementioned challenges together make it difficult to address UQ tasks for dynamic multiphase flow models using traditional surrogate methods and call for innovative solutions.

Deep neural networks (DNNs) have achieved an increased popularity in many communities such as computer vision~\citep{badrinarayanan2017,he2016}, dimensionality reduction~\citep{chan2017,LALOY2017387,LALOY2018}, and surrogate modeling~\citep{tripathy2018,zhu2018} due to their robustness and generalization property. The basic idea of DNNs for surrogate modeling is that one can approximate underlying functions through a hierarchy of hidden layers of increasing complexity. Deep neural networks tackle the curse of dimensionality though a series of inherent nonlinear projections of the input into exploitable low-dimensional latent spaces. They also scale well with the dimensionality due to the minibatch approach used when training the network~\citep{Goodfellow-et-al-2016}. 

Two successful applications of DNNs in UQ of elliptic stochastic PDEs with high-dimensional random inputs were reported recently~\citep{zhu2018,tripathy2018}. In~\citet{tripathy2018}, the DNNs were used to approximate some scalar quantity of interest and were capable of making predictions given input fields at arbitrary correlation lengths.~\cite{zhu2018}   transformed the surrogate modeling task to an image-to-image regression task by using a convolutional encoder-decoder network. The encoder is used to extract coarse features from the high-dimensional input field which are subsequently used by the decoder to reconstruct the output fields. The high-dimensional input and output fields were treated as images and a dense fully convolutional network architecture was used to alleviate the `big data' requirement of deep learning. Both the encoder and decoder were composed of fully convolutional neural networks (CNNs), which are well suited for image processing as they explicitly consider the spatial structure of the data~\citep{Goodfellow-et-al-2016,GU2018,LALOY2017387,LALOY2018}. This method showed promising computational efficiency in UQ for problems with up to $4225$ input dimensions~\citep{zhu2018}. 

In this paper, we extend the dense fully convolutional encoder-decoder network architecture presented in~\citet{zhu2018} to propose a deep convolutional encoder-decoder network method to efficiently address the UQ problem for dynamic multiphase flow models with high-dimensional inputs and outputs. We adopt an image-to-image regression strategy. The employed network architecture  substantially strengthens the information propagation through the network, providing an efficient way to obtain an accurate surrogate to our dynamic system. To accurately characterize the discontinuous saturation front,  we introduce a two-stage network training strategy combining a regression loss with a segmentation loss. The intuition behind introducing the segmentation loss is inspired by the fact that for a sharp saturation front, it is easy to classify if a spatial grid point has a non-zero saturation. A binary ($0$ or $1$) image is added as an extra output of the system capturing this information. The segmentation loss is mostly caused by the mismatch in regions around the saturation front. We will show that this strategy can signiﬁcantly improve the approximation of the discontinuous saturation front especially when the training data is limited. To address the dynamic nature of the problem, time will be treated as an extra network input. We will demonstrate that our model, trained with  outputs at only a limited number of time instances, is capable of accurately and efficiently predicting the high-dimensional output fields at arbitrary time instances. The overall integrated methodology is applied to a highly nonlinear multiphase flow  model in geological carbon storage. The developed algorithm is non-intrusive and can be employed to a wide range of stochastic problems.

The rest of the paper is organized as follows. In section~\ref{section:PF}, we introduce a multiphase flow model as relates to geological carbon storage   and define the UQ problem of interest for a high-dimensional input. In section~\ref{section:method}, we present our proposed deep learning method. In section~\ref{section:exp}, we evaluate the performance of the model in uncertainty quantification tasks. 
Finally, in section~\ref{section:conc}, a summary is provided together with a discussion on potential extensions.

\section{Problem Formulation}\label{section:PF}

Our focus is the development of a surrogate model for dynamic multiphase problems that we can use to efficiently propagate the   uncertainty associated with heterogeneous media properties (e.g., permeability and porosity) to system responses. We consider multiphase (CO$_2$ as gas and water as liquid) flow and multi-component (H$_2$O, NaCl, and CO$_2$) transport equations as the GCS model of interest. GCS is a promising strategy to reduce the emission of CO$_2$ into the atmosphere by injecting it into deep saline aquifers for permanent storage~\citep{benson2008}. The injection of supercritical CO$_2$ into a deep geological formation leads to pressure buildup and CO$_2$  plume evolution. Characterizing the pressure distribution and the CO$_2$ plume migration over time is crucial for predicting the fate of injected CO$_2$ and for risk assessment of GCS projects~\citep{BIRKHOLZER2015,Celia2015,Cihan2013,LiYJ2017}. For example, the increase of pressure due to CO$_2$ injection may break the integrity of the caprock, leading to the escape of injected CO$_2$ and contamination of upper groundwater aquifers.

\subsection{Governing Equations}\label{GE}

In GCS modeling, for each component $\kappa$, we have a mass conversation equation~\citep{pruess1999}
 \begin{linenomath*}
\begin{equation}\label{spde}
    \frac{\partial{M^\kappa}}{\partial{t}}=-\nabla\cdot{\textbf{\emph F}}^\kappa+q^\kappa,
\end{equation}
\end{linenomath*}
\noindent where $M$ is the mass accumulation term, \textbf{\emph F} is the sum of mass flux over phases, and $q$ denotes sinks and sources. The total mass of component $\kappa$ is obtained by summing over  phases
 \begin{linenomath*}
\begin{equation}
    M^\kappa={\phi}\sum_{\beta}S_{\beta}\rho_{\beta}X_{\beta}^{\kappa},
\end{equation}
\end{linenomath*}
\noindent where $\phi$ is the porosity, $S_{\beta}$ is the saturation of phase $\beta$, $\rho_{\beta}$ is the density of phase $\beta$, and $X_{\beta}^{\kappa}$ is the mass fraction of component $\kappa$ present in phase $\beta$. In the GCS model, we do not consider the modular diffusion and hydrodynamic dispersion, thus mass transport only occurs by  advection. Similarly, the advective mass flux is a sum over phases,
 \begin{linenomath*}
\begin{equation}
    \textbf{\emph{F}}^{\kappa}\mid_{\rm{adv}}=\sum_{\beta}X_{\beta}^{\kappa}\textbf{\emph{F}}_{\beta},
\end{equation}
\end{linenomath*}
\noindent and individual phase fluxes are given by a multiphase version of Darcy's flow equation:
 \begin{linenomath*}
\begin{equation}
    \textbf{\emph{F}}_\beta=-k\frac{k_{\rm{r},\beta}\rho_\beta}{\mu_\beta}(\nabla{P_\beta}-\rho_{\beta}\textbf{\emph g}).
\end{equation}
\end{linenomath*}
Here, $k$ is the absolute permeability, $k_{\rm r,\beta}$ is the relative permeability for phase $\beta$, $\mu_{\beta}$ is the viscosity, and
 \begin{linenomath*}
\begin{equation}
    P_\beta=P+P_{\rm{c},\beta},
\end{equation}
\end{linenomath*}
\noindent is the fluid pressure in phase $\beta$, which is the sum of the pressure $P$ of a reference phase (usually taken to be the gas phase), and the capillary pressure $P_{\rm{c},\beta}$ ($\le{0}$). $\textbf{\emph g}$ is the gravitational acceleration vector. 

\subsection{Uncertainty Quantification}
\label{sec:UQ}
When the coefficients of the PDE in equation~(\ref{spde}) are treated as random fields, the solution is no longer deterministic and the underlying equation becomes a stochastic PDE. Consequently, it is indispensable to characterize the effect of parametric uncertainty on the solution by means of, e.g., estimating statistical moments of the output responses. More specifically, let $\textbf{\emph y}=\boldsymbol{\eta} \big(\textbf{\emph s},t,\textbf{\emph x}(\textbf{\emph s})\big)$ denote the solution of equation~(\ref{spde}), i.e., $\textbf{\emph{y}}\in{\mathbb{R}^{d_y}}$, at  spatial location $\textbf{\emph{s}}\in \mathcal{S}\subset{\mathbb{R}^{d_s}}$ $(d_s=1,2,3)$, where $\mathcal S$ is the location index set, and at the time instant $t\in \mathcal T \subset \mathbb R $ with one realization $\textbf {\emph x}(\textbf {\emph s}) \in \mathbb R^{d_x}$ of the random input fields $\{\textbf{\emph x}(\textbf{\emph s}), \textbf{\emph s}\in \mathcal S\}$. In practice, the numerical simulation is performed over a set of spatial grid locations $\mathcal{S}=\{\textbf{\emph s}_1,\ldots,\textbf{\emph s}_{n_s}\}$ (e.g. grid blocks in finite differences approximations) in the physical domain. In this case, the random field $\textbf{\emph x}$ is discretized over the $n_s$ grid points, which results in a high-dimensional vector, denoted as $\textbf x$, where $\textbf x\in \mathcal X \subset \mathbb R^{d_x n_s}$.  The corresponding response $\textbf{\emph y}$ is solved over $\mathcal{S}$, thus can be represented as a vector $\textbf y=\boldsymbol{\eta}(\mathbf{x}, t)\in \mathcal Y \subset \mathbb R^{d_y n_s}$, where  $t = t_1, \cdots, t_{n_t}$ are the time instances considered. 

The objective of the classical UQ problem is the estimation of the first two statistical moments (mean and variance) of the output $\boldsymbol{\eta}(\textbf x,t)$, i.e., the mean,
 \begin{linenomath*}
\begin{equation}
\label{mean}
    \boldsymbol{\mu}=\int_{\Omega}\boldsymbol{\eta}(\textbf x,t) p(\textbf x) d\textbf x,
\end{equation}
\end{linenomath*}
\noindent and the variance,
 \begin{linenomath*}
\begin{equation}
\label{variance}
    \boldsymbol{\sigma}^2=\int_{\Omega}\big(\boldsymbol{\eta}(\textbf x,t)-\boldsymbol{\mu} \big)^2 p(\textbf x) d\textbf x,
\end{equation}
\end{linenomath*}
\noindent where $\Omega$ is the sample space and $p(\textbf x)$ is the probability distribution of the random input field. We are also interested in the probability density function (PDF) of $\boldsymbol{\eta}(\textbf x,t)$. The MC methods are commonly used to obtain the numerical approximations of the moments and PDF. Unfortunately, the MC method is computationally intensive because of its slow convergence rate. In order to accelerate the convergence of MC simulation (i.e., using fewer model runs), we propose the development of a deep convolutional encoder-decoder neural network surrogate that replaces the GCS model when performing MC simulation.

\section{Methods}\label{section:method}

\subsection{Deep Convolutional Neural Networks}

Neural networks approximate the input-output relationship $\boldsymbol{\eta}:\mathcal X\to\mathcal Y$ through a hierarchy of layers which are combinations of a set of neurons, as represented by
 \begin{linenomath*}
\begin{equation}
    h(\textbf x)=f(\textbf x^\top\boldsymbol\omega+b),
\end{equation}
\end{linenomath*}
\noindent where $h(\cdot)$ is the output of the neuron, $f(\cdot)$ is a nonlinear activation function, $\boldsymbol\omega$ and $b$ are a weight vector of the same dimension as $\textbf x$ and a scalar bias, respectively. Popular choices for $h(\cdot)$ include the rectified linear unit (ReLU), sigmoid function or the hyperbolic tangent function~\citep{Goodfellow-et-al-2016}. In this work, the bias is not considered, thus $h(\textbf x)=f(\textbf x^\top\boldsymbol\omega)$.

In neural networks, the neurons are organized in layers. A deep neural network is simply a neural network with multiple intermediate (hidden) layers. The output of the $l^{th}$ layer of the network is given by:
 \begin{linenomath*}
\begin{equation}
    \textbf z^{(l)}=\boldsymbol h_l(\textbf z^{(l-1)})=f_l(\textbf W^{(l)}\textbf z^{(l-1)}),\quad \forall l\in\{1,\ldots,L_{NN}\},
\end{equation}
\end{linenomath*}
\noindent where $f_l$ denotes the activation function of the $l^{th}$ layer, $\textbf W^{(l)}\in \mathbb R^{d_ld_{l-1}}$ is the weight matrix,  $d_l$ is the number of neurons in the $l^{th}$ layer, and $L_{NN}$ is the number of layers of the neural network. Here, $\textbf z^0$ is taken as the input $\textbf x$.  

Fully connected neural networks may lead to an extremely large number of network parameters. Convolutional  neural networks   are commonly used  to greatly reduce the parameters being learnt~\citep{Goodfellow-et-al-2016,GU2018} since they lead to sparse connectivity and parameter sharing. They are  particularly suited for image processing because they exploit the spatially-local correlation of the data by enforcing a local connectivity pattern between neurons of adjacent layers~\citep{Goodfellow-et-al-2016,GU2018,LALOY2017387,LALOY2018}. A convolutional layer is composed of a series of convolution kernels which are used to compute the feature maps that are essentially matrices. Specifically, when the input is a 2-D image, a convolutional layer, $\textbf{\emph h}$, is obtained by employing a series of $q=1,\ldots,N_{\rm{out}}$ filters $\boldsymbol\omega^q \in\mathbb R^{k'_i\times k'_j}$, where $k'$ is referred to as kernel size, to evolve an input pixel $x_{u,v}$ to obtain the feature value $h_{u,v}^q(x_{u,v})$ at location $(u,v)$ as
 \begin{linenomath*}
\begin{equation}
    h_{u,v}^q(x_{u,v})=f\left(\sum_{i=1}^{k'_i}\sum_{j=1}^{k'_j}\omega_{i,j}^q x_{u+i,v+j}\right).
\end{equation}
\end{linenomath*}
This results in a convolutional layer $\textbf{\emph h}$ consisting of $N_{\rm{out}}$ feature maps, i.e., $\{\textbf{\emph h}^q, q=0,\ldots,N_{\rm{out}}\}$. Two other important parameters for the convolutional layer are the stride, $s$, and zero padding, $p$, which determine the distance between two successive moves of the filter and the padding of the borders of the input image with zeros for size preservation, respectively. For square inputs, square kernel size (i.e., $k'_i=k'_j=k'$), same strides and zero padding along both axes, the output feature map size  can be calculated by~\citep{dumoulin2016}
 \begin{linenomath*}
\begin{equation}
    S_{\rm{out}}=\left \lfloor \frac{S_{\rm{in}}+2p-k'}{s} \right \rfloor+1,
\end{equation}
\end{linenomath*}
\noindent where $S_{\rm{in}}$ is the input feature map size and $\left\lfloor\cdot\right\rfloor$ denotes the floor function. 

\subsection{Deep Convolutional Endoder-Decoder Network for Image-to-Image Regression}

We employ a dense fully convolutional encoder-decoder architecture to formulate the deep convolutional encoder-decoder network method. The adopted network architecture has shown a promising performance in efficiently handling the mapping between high-dimensional inputs and outputs of a steady-state Darcy flow model~\citep{zhu2018}. The surrogate modeling task is transformed into an image-to-image regression problem by employing an encoder-decoder network structure based on fully convolutional neural networks. In the following subsections, the image-to-image regression strategy and a state-of-art   architecture based on `dense blocks' are briefly reviewed for completeness of the presentation. For more details, the interested reader can  refer to~\citet{zhu2018}. 

\subsubsection{Surrogate Modeling as Image-to-Image Regression}

In section~\ref{sec:UQ}, we described the model input-output relationship as a mapping of the form:
 \begin{linenomath*}
\begin{equation}
    \boldsymbol{\eta}:\mathcal X\to\mathcal Y,
\end{equation}
\end{linenomath*}
\noindent where $\mathcal X\subset\mathbb R^{d_xn_s}$ and $\mathcal Y\subset\mathbb R^{d_yn_s}$, $n_s$ is the number of grid blocks, $d_x$ and $d_y$ denote the dimension of the input and output at one grid block, respectively. Assume that the PDEs defined in equation~(\ref{spde}) are solved over 2-D regular grids of size $H\times W$, where $H$ and $W$ denote the number of grid blocks in the two axes of the spatial domain (height and width), and $n_s=H\times W$. We can reorganize the input $\textbf x\in\mathbb R^{d_xn_s}$ and output $\textbf y\in\mathbb R^{d_yn_s}$ as image-like data, i.e., $\textbf x\in\mathbb R^{d_x\times H\times W}$ and output $\textbf y\in\mathbb R^{d_y\times H\times W}$. Therefore, the surrogate modeling problem is transformed to an image-to-image regression problem, with the regression function as 
 \begin{linenomath*}
\begin{equation}
    \boldsymbol{\eta}:\mathbb R^{d_x\times H\times W}\to\mathbb R^{d_y\times H\times W}.
\end{equation}
\end{linenomath*}
\noindent It is straightforward to generalize to the 3-D spatial domain by adding an extra depth axis to the images, i.e., $\textbf x\in\mathbb R^{d_x\times D\times H\times W}$ and  $\textbf y\in\mathbb R^{d_y\times D\times H\times W}$. The image regression problem here becomes a pixel-wise prediction task, i.e., predicting the outputs at each grid block. This is solved by employing a convolutional  encoder-decoder neural network architecture which shows a good performance in pixel-wise segmentation~\citep{badrinarayanan2017} and prediction~\citep{zhu2018}. 

\subsubsection{Dense Convolutional Encoder-Decoder Networks}

The intuition behind the encoder-decoder architecture for the regression between two high-dimensional images is to go though a coarse-refine process, i.e., to extract high-level coarse features from the input images using an encoder, and then refine the coarse features to output images through a decoder. Since normal deep neural networks are data-intensive, a densely connected convolutional neural network architecture called dense block~\citep{huang2017} is employed in the encoder-decoder architecture to enhance information (features) propagation through the network and reduce network parameters, thus to construct accurate surrogate models with limited training data. 

\begin{figure}[h!]
\centering
\includegraphics[width=0.9\textwidth]{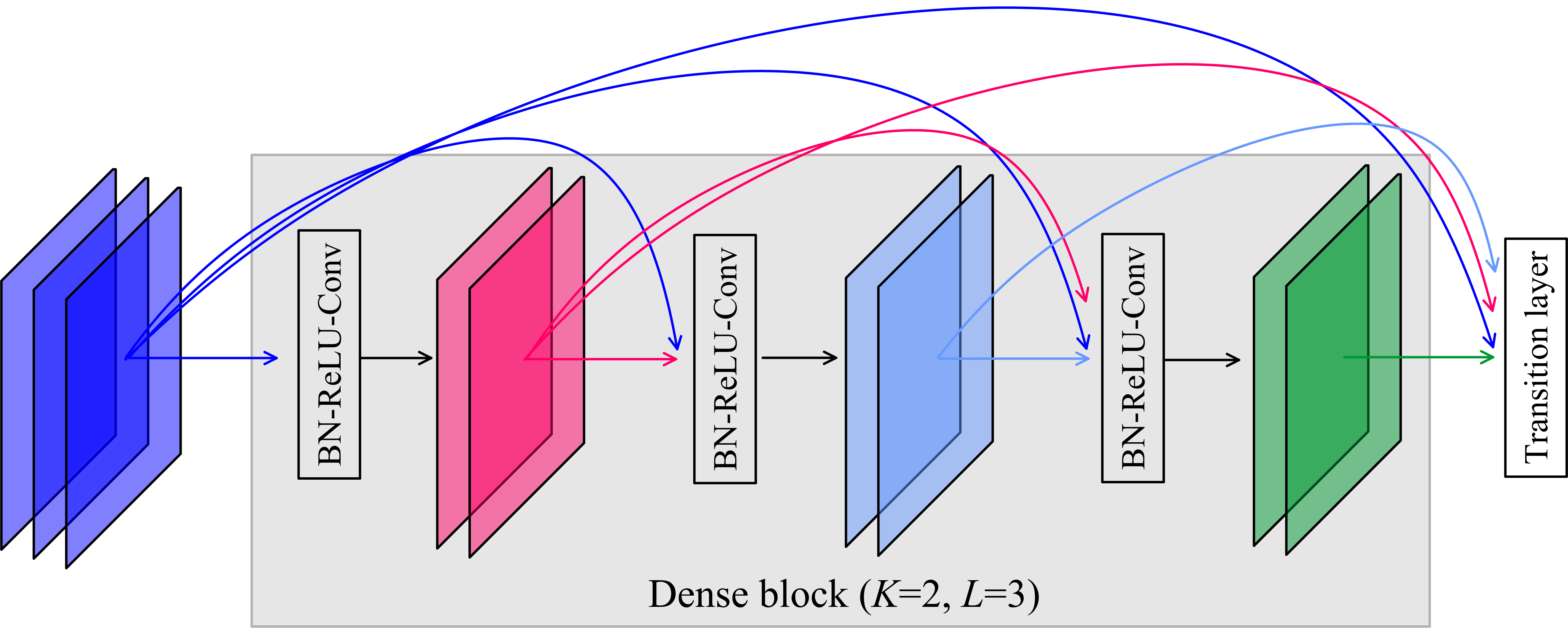}
    \caption{A dense block with $L$=3 layers and growth rate $K$=2. The input to a layer is the concatenation of the output and input feature maps of the previous layers. Each layer contains three different consecutive operations: BN, followed by ReLU, and Conv.}
\label{denseblock}
\end{figure}

In the dense block, the outputs of each layer are connected with all successor layers. More specifically, the output $\textbf z^{(l)}$    of the $l^{th}$ layer  receives all the feature maps from the previous layers as input:
\begin{linenomath*}
\begin{equation}
    \textbf z^{(l)}=\boldsymbol h_l([\textbf z^{(l-1)},\ldots,\textbf z^0]),
\end{equation}
\end{linenomath*}
where $[\textbf z^{(l-1)},\ldots,\textbf z^0]$ denote the concatenated output features from layers $0$ to $l-1$. The dense block contains two design parameters determining its structure, namely the number of the layers   $L$ within the block and the growth rate $K$, which is the number of output features of each single layer. Figure~\ref{denseblock} shows the conceptual diagram of a dense block with $L=3$ and $K=2$. It contains three different consecutive operations: batch normalization (BN)~\citep{ioffe2015batch}, followed by ReLU~\citep{Goodfellow-et-al-2016} and convolution (Conv)~\citep{al2016theano}.  

\begin{figure}[h!]
\centering
\includegraphics[width=0.55\textwidth]{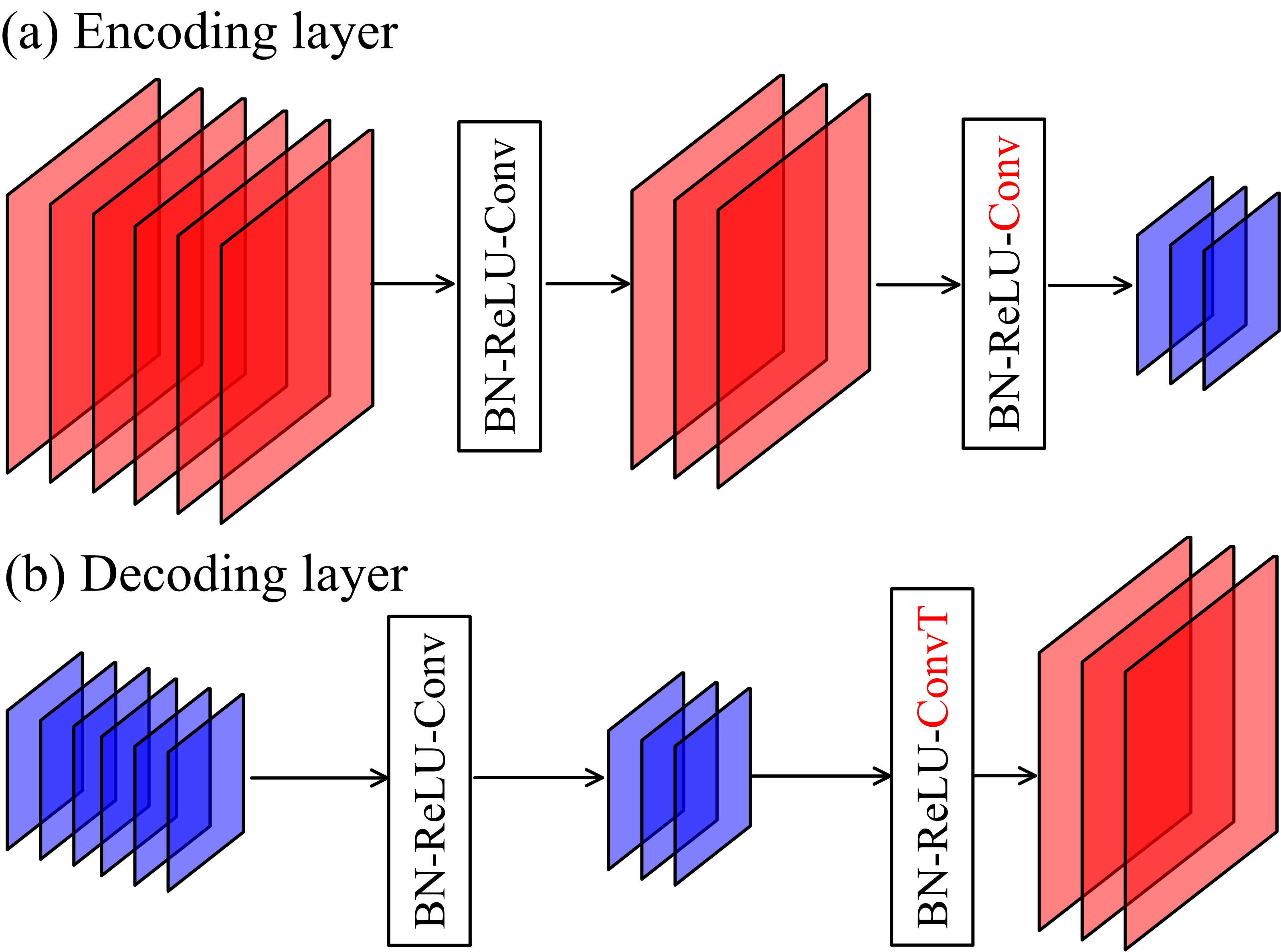}
    \caption{(a) An encoding layer and (b) a decoding layer, both of which contain two convolutions (Convs). The first Conv reduces the number of feature maps while keeps their size the same; the second Conv changes the size of the feature maps but keeps their number the same. The main difference between (a) and (b) is the type of the second Conv layer, which is  Conv for downsampling and transposed Conv (ConvT) for upsampling, respectively.}
\label{EnDecode}
\end{figure}

The output feature maps of a dense block are then fed into a transition layer.  The transition layer, which is referred to as encoding layer in the encoder and as decoding layer in the decoder, is placed between two adjacent dense blocks to avoid feature maps explosion and to change the size of feature maps during the coarse-refine process. It halves the size of feature maps in the encoder (i.e., downsampling), while doubles the size in the decoder (i.e., upsampling). Both the encoding and decoding layers reduce the number of feature maps. An illustration of the encoding and decoding layers is given in Figure~\ref{EnDecode}. 

\begin{figure}[h!]
\centering
\includegraphics[width=0.7\textwidth]{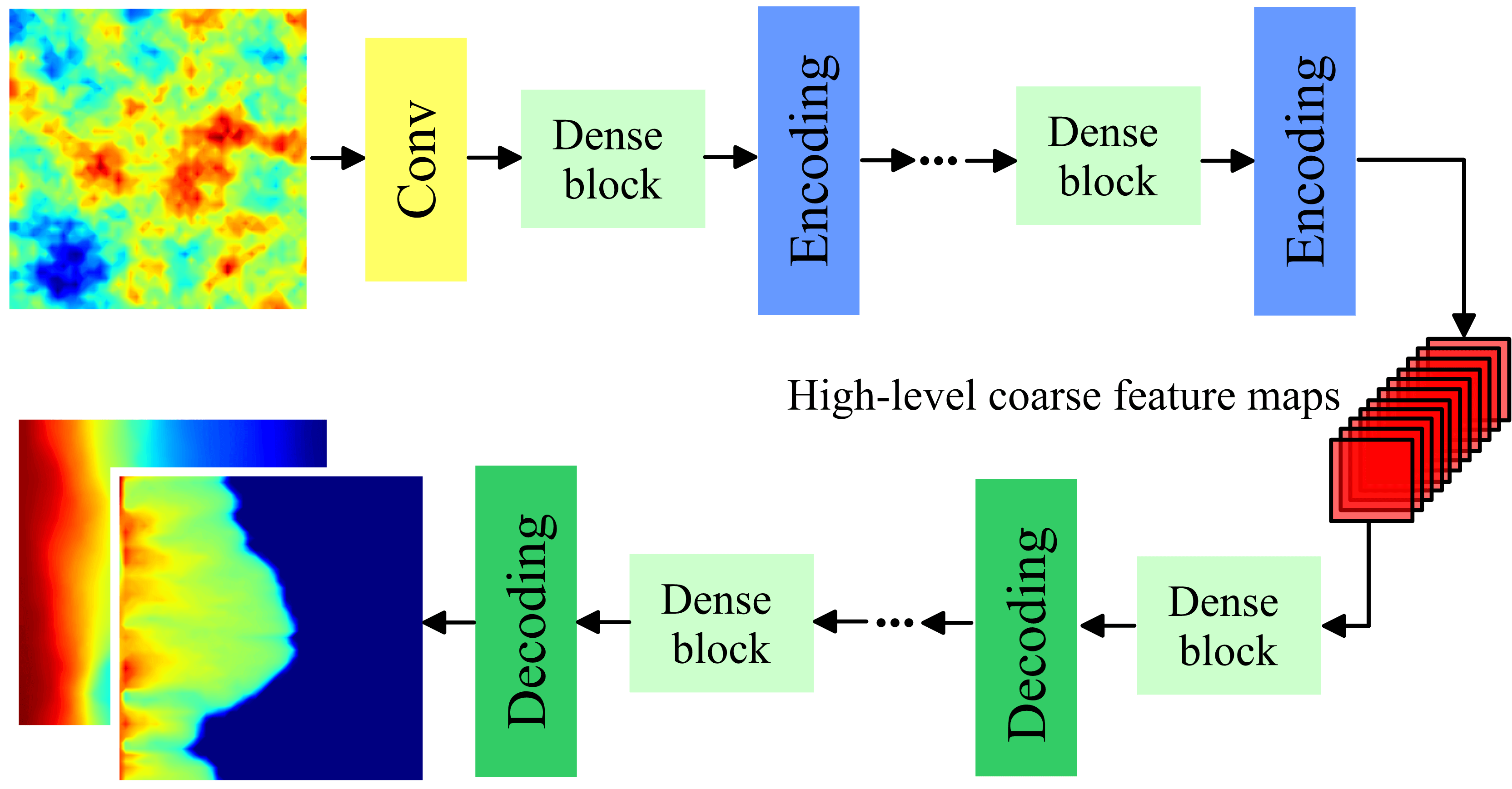}
    \caption{Illustration of the network architecture of the deep convolutional encoder-decoder network. The encoder operates on the input images to extract a set of high-level coarse feature maps (red boxes), which are then fed into the decoder to eventually construct the output images.}
\label{DCEDN}
\end{figure}

The network integrated architecture employed is depicted in Figure~\ref{DCEDN}. The network is fully convolutional without any fully connected layers. Such architectures show better performance in pixel-wise classification compared to convolutional neural networks containing fully connected layers~\citep{long2015}. In the encoding path, the input image is fed into the first convolution layer. The extracted feature maps are then passed through an alternating cascade of dense blocks and encoding layers. The last encoding layer outputs high-level coarse feature maps, as shown with red boxes, which are subsequently fed into the decoder. The decoder is composed of an alternation of dense blocks and decoding layers, with the last decoding layer leading to the output images. 

 \subsection{Deep Convolutional Encoder-Decoder Network for Approximating Time-Dependent Outputs}

For dynamic systems, it is important to   develop a surrogate that allows prediction of responses at arbitrary time instances. It is certainly computationally inefficient to construct independently surrogate models for outputs at all   time instances of interest. A computationally attractive alternative is to treat the time $t$ as an input to the surrogate model and train the surrogate model with the outputs at only a limited number of time instances. Then the surrogate can potentially make predictions at time instances different from those used in the training data. 

The surrogate for dynamic systems is
\begin{linenomath*}
\begin{equation*}
    \hat{\mathbf{y}}^{i,j} = \mathbf{f}(\mathbf{x}^i, t_j; \bm{\theta}),
\end{equation*}
\end{linenomath*}
where $\hat{\mathbf{y}}^{i,j} \in \mathbb{R}^{d_y \times H \times W}$ is the model prediction at the input $\mathbf{x}^i \in \mathbb{R}^{d_x \times H \times W}$ and time $t_j$, and $\bm{\theta}$ denotes all the network parameters. The training data is organized as $\{(\mathbf{x}^i, t_j; \mathbf{y}^{i,j})\}_{i=1,j=1}^{N, n_t} = \{(\tilde{\mathbf{x}}^m, \mathbf{y}^m)\}_{m=1}^{Nn_t} = (\widetilde{\mathbf{X}}, \mathbf{Y})$, where $\mathbf{y}^{i,j}$ denotes the simulation output, $\tilde{\mathbf{x}}^m = (\mathbf{x}^i, t_j)$ and $m$ is a re-index for $(i, j)$. Also, $N$ and $n_t$ denote the total number of model runs and time instances when simulation data is stored in each run, respectively.

The time instance $t_j$ is broadcast as an extra feature map and concatenated with the latent feature maps (as shown with red boxes in Figure~\ref{DCEDN}) extracted from the input $\mathbf{x}^i$ since they independently impact the output $\mathbf{y}^{i,j}$.

\subsection{Two-Stage Network Training Combining  Regression and Segmentation Losses}\label{2stepTrain}

Training the surrogate amounts to optimize the network parameters $\mathbf{\bm{\theta}}$ using training data with respect to certain loss function. The mean squared error (MSE) loss for the regression problem is defined as
\begin{linenomath*}
\begin{equation}
    L_{\text{MSE}}(\bm\theta; \widetilde{\mathbf{X}}, \mathbf{Y}) = \frac{1}{N  n_t}\sum_{i=1}^{N} \sum_{j=1}^{n_t} l_{\text{MSE}}(\bm\theta; \mathbf{x}^i, t_j, \mathbf{y}^{i, j}), 
\label{eq:MSE}
\end{equation}
\end{linenomath*}
where 
\begin{linenomath*}
\begin{equation}
l_{\text{MSE}}(\bm\theta; \mathbf{x}^i, t_j, \mathbf{y}^{i, j}) = ||\mathbf{f}(\mathbf{x}^i, t_j; \bm\theta) - \mathbf{y}^{i, j} ||_2^2.
\end{equation}
\end{linenomath*}

\begin{figure}[h!]
	\centering
	\includegraphics[width=0.6\textwidth]{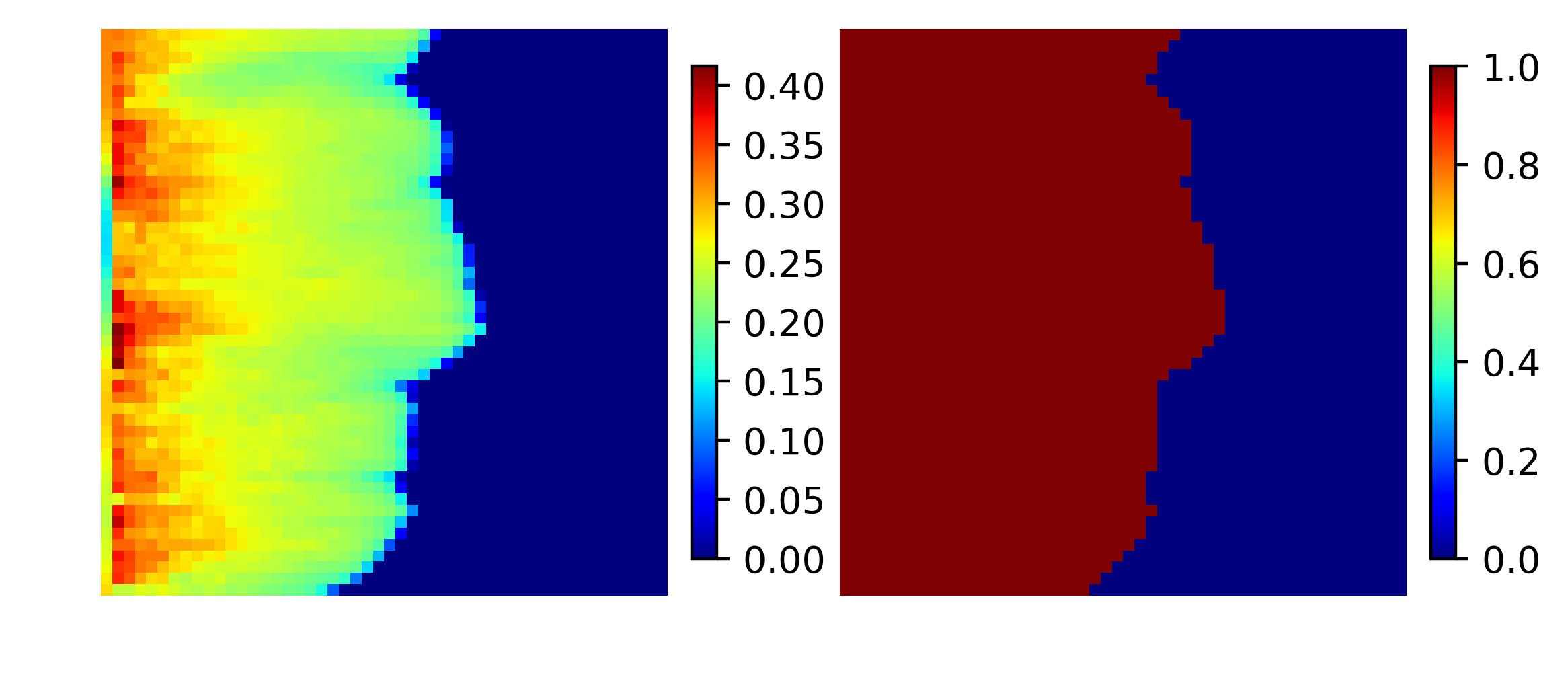}
	\caption{One example of the saturation field (left) and the corresponding binary image with the response value at pixels being $0$ or $1$.}
	\label{Sg_binarized}
\end{figure}

Even though deep neural networks have a robust capacity to model complex output responses, accurately approximating the discontinuous saturation front in our multi-phase flow system remains a challenge. We address this problem by augmenting the original regression task with image segmentation for the output saturation field, which separates the area with saturation greater than zero from the area with zero saturation. The training data for segmentation is obtained by binarizing the saturation field. More specifically,
we define the following indicator function pixel-wise:
\begin{linenomath*}
\begin{equation}\label{Closs}
   \zeta(Sg)=
   \begin{cases}
      0,& Sg=0\\
      1,& Sg>0
   \end{cases}
   ,
\end{equation}
\end{linenomath*}
where $Sg$ is the saturation. Given a saturation field, we can obtain a corresponding binary image with the response value at each pixel being $0$ or $1$. Figure~\ref{Sg_binarized} shows an example of the saturation field and the corresponding binary image obtained by applying equation~(\ref{Closs}). The discontinuous saturation front in Figure~\ref{Sg_binarized}a is clearly reflected by the boundary between $0$ and $1$ in Figure~\ref{Sg_binarized}b. 

The binary saturation field $\bm\zeta\in \mathbb{R}^{1 \times H \times W}$ is treated as an extra output field of the forward model. Now the surrogate output becomes 
\begin{linenomath*}
\begin{equation}
    [\hat{\mathbf{y}}, \hat{\bm\zeta}] = \mathbf{f}(\mathbf{x}, t; \bm\theta) \in \mathbb{R}^{(d_y + 1) \times H \times W}.
\end{equation} 
\end{linenomath*}
The segmentation output $\hat{\bm\zeta}$ is expected to capture the saturation front very well and thus carries this statistical strength to the original regression task that we are interested in.

We use the standard binary cross entropy (BCE) loss for this two-class segmentation task with training data $(\widetilde{\mathbf{X}}, \mathbf{Z}) = \{\mathbf{x}^i, t_j; \bm\zeta^{i, j} \}_{i=1, j=1}^{N, n_t}$
\begin{linenomath*}
\begin{equation}
    L_{\text{BCE}}(\bm\theta; \widetilde{\mathbf{X}}, \mathbf{Z})=\frac{1}{N n_t}\sum_{i=1}^{N} \sum_{j=1}^{n_t} l_{\text{BCE}}(\bm\theta; \mathbf{x}^i, t_j, \bm\zeta^{i,j}),
\end{equation}
\end{linenomath*}
where 
\begin{linenomath*}
\begin{equation}
\label{eq:bce}
    l_{\text{BCE}}(\bm\theta; \mathbf{x}^i, t_j,\bm\zeta^{i,j})=\frac{1}{n_s}\sum_{k=1}^{n_s}\bm\zeta_k(\mathbf x^i)\log \hat{\bm\zeta}_k (\mathbf x^i, t_j; \bm\theta)+(1-\bm\zeta_k(\mathbf x^i))\log \left(1-\hat{\bm\zeta}_k (\mathbf x^i, t_j; \bm\theta)\right),
\end{equation}
\end{linenomath*}
with $n_s$ being the number of pixels in the binary image.

In network training, the stochastic gradient descent (SGD) algorithm  is used as the optimizer for parameter learning~\citep{Goodfellow-et-al-2016}. Let $J(\boldsymbol\theta)$ denote the loss function. In our model, the network is trained in a two-stage manner for each epoch:

\noindent(1) Train the network using the regularized MSE loss, i.e.,
\begin{linenomath*}
\begin{equation}\label{MSEloss}
    J_{\text{MSE}}(\boldsymbol\theta)=L_{\text{MSE}}(\bm\theta; \widetilde{\mathbf{X}}, \mathbf{Y}) + \frac{\alpha}{2} \bm\theta^\top\bm\theta,
\end{equation}
\end{linenomath*}
where $\alpha$ is the regularization coefficient, also called weight decay.

\noindent(2) Using the optimized parameter values in Step~($1$) as the starting point, optimize the network parameters based on a combination of the MSE and BCE losses, i.e.,
\begin{linenomath*}
\begin{equation}\label{eq:Hybridloss}
    J_{\text{MSE,BCE}}(\bm\theta)=L_{\text{MSE}}(\bm\theta; \widetilde{\mathbf{X}}, \mathbf{Y}) + w L_{\text{BCE}}(\bm\theta; \widetilde{\mathbf{X}}, \mathbf{Z}) + \frac{\alpha}{2} \bm\theta^\top\bm\theta,
\end{equation}
\end{linenomath*}
where $w$ is a weight balancing the two losses. 

In the first stage, the network is trained aiming to accurately approximate the output fields as  a  traditional network will do in regression tasks. The second stage aims to refine the approximation around the discontinuous saturation front. As mentioned above, the segmentation loss is   attributed to the  mismatch in the regions around the discontinuous front, thus minimization of this loss can promote a further improvement in the approximation accuracy in those regions. 

The SGD algorithms require computing the gradient of the loss $J(\boldsymbol\theta)$ with respect to $\boldsymbol\theta$. Taking the MSE loss $J_{\text{MSE}}(\boldsymbol\theta)$ as an example, we can write:
\begin{linenomath*}
\begin{equation}
    -\nabla_{\boldsymbol\theta}J_{\text{MSE}}(\boldsymbol\theta)=-\frac{1}{Nn_t}\sum_{m=1}^{Nn_t}\nabla_{\boldsymbol\theta}L_{\text{MSE}}(\tilde{\mathbf{ x}}^m,\mathbf y^m,\boldsymbol\theta).
\end{equation}
\end{linenomath*}
The computational cost of this operation may be very high when $Nn_t$ is large. In SGD, a minibatch strategy is used to reduce the computational burden, in which only a minibatch of samples uniformly drawn from the training set ($M< Nn_t$) are used~\citep{Goodfellow-et-al-2016}. Then the estimate of the gradient is given as
 \begin{linenomath*}
\begin{equation}
    \boldsymbol g=-\frac{1}{M}\sum_{m=1}^{M}\nabla_{\boldsymbol\theta}L_{\text{MSE}}(\tilde{\mathbf{ x}}^m,\mathbf y^m,\boldsymbol\theta).
\end{equation}
\end{linenomath*}
\noindent The SGD then follows the estimated gradient downhill:
 \begin{linenomath*}
\begin{equation}
    \boldsymbol\theta\leftarrow\boldsymbol\theta+\varepsilon\boldsymbol g,
\end{equation}
\end{linenomath*}
where $\varepsilon$ is the learning rate. Various SGD algorithms are available. The widely used Adam algorithm~\citep{Kingma2014} is adopted in this work.

The two-stage training strategy is summarized in Algorithm~\ref{2stage}. Briefly speaking, in each epoch, all minibatches of the training sample set are utilized to optimize the network parameters. For each minibatch, the network parameters $\boldsymbol\theta$ are updated twice: firstly updated using the gradient of $J_{\text{MSE}}(\boldsymbol\theta)$ (i.e., lines $4$ and $5$ in Algorithm~\ref{2stage}); and then updated using the gradient of $J_{\text{MSE,BCE}}(\boldsymbol\theta)$ (i.e., lines $6$  and $7$ in Algorithm~\ref{2stage}). The obtained values are then used as the starting point when using the next minibatch to update the parameters. In section~\ref{section:exp}, we will compare the performance of the network trained using the two-stage training strategy (i.e., Algorithm~\ref{2stage}) with the network based solely on the traditional MSE loss (i.e., the stage one in Algorithm~\ref{2stage}). This study will illustrate the effectiveness of the proposed two-stage training strategy in handling discontinuous outputs.

\begin{algorithm}
\caption{Two-stage training strategy to optimize the network parameters $\boldsymbol\theta$. The default network configurations is shown in Figure~\ref{DCEDN_used}, $\alpha=0.0005$, $w=0.01$, $M=100$, $\varepsilon=0.001$, $\beta=(0.9, 0.999)$}
\label{2stage}
\begin{algorithmic}[1]
\Require Encoder-Decoder network configurations, weight decay $\alpha$, segmentation weight $w$, mini-batch size $M$, Adam hyperparameters $\varepsilon, \beta$.
\State $\boldsymbol\theta \gets \boldsymbol\theta_0$ \Comment{Initialization}
\For{number of epochs}
  \For{\textbf{each} minibatch $\{(\tilde{\mathbf{x}}^m, \mathbf{y}^m, \bm\zeta^m\}_{m=1}^M$ of the training sample set}
     \State $\boldsymbol g \gets -\nabla_{\boldsymbol\theta} \Big[\frac{1}{M}\sum_{m=1}^M l_{\text{MSE}}(\bm\theta; \tilde{\mathbf{x}}^m, \mathbf{y}^m) +  \frac{\alpha}{2} \bm\theta^\top\bm\theta \Big]$ \Comment{Stage one}
     \State $\boldsymbol\theta \gets \text{Adam}(\bm\theta, \mathbf{g}, \varepsilon, \beta)$ \Comment{$\varepsilon$ is subject to learning rate scheduling}
     \State $\boldsymbol g \gets -\nabla_{\bm\theta} \big[ \frac{1}{M} \sum_{m=1}^M ( l_{\text{MSE}}(\bm\theta; \tilde{\mathbf{x}}^m, \mathbf{y}^m) + w l_{\text{BCE}}(\bm\theta; \tilde{\mathbf{x}}^m, \bm\zeta^m) ) + \frac{\alpha}{2} \bm\theta^\top\bm\theta \Big]$ \Comment{Stage two}
     \State $\boldsymbol\theta \gets \text{Adam}(\bm\theta, \mathbf{g}, \varepsilon, \beta)$
  \EndFor
  \State \textbf{end for}
\EndFor
\State \textbf{end for}
\State \textbf{return} $\boldsymbol\theta$\Comment{The optimized $\boldsymbol\theta$}
\end{algorithmic}
\end{algorithm}

\section{Numerical Experiments}\label{section:exp}
\subsection{Experiment Setting}
\subsubsection{GCS Multiphase Flow Model}

\begin{figure}[h!]
\centering
\includegraphics[width=0.6\textwidth]{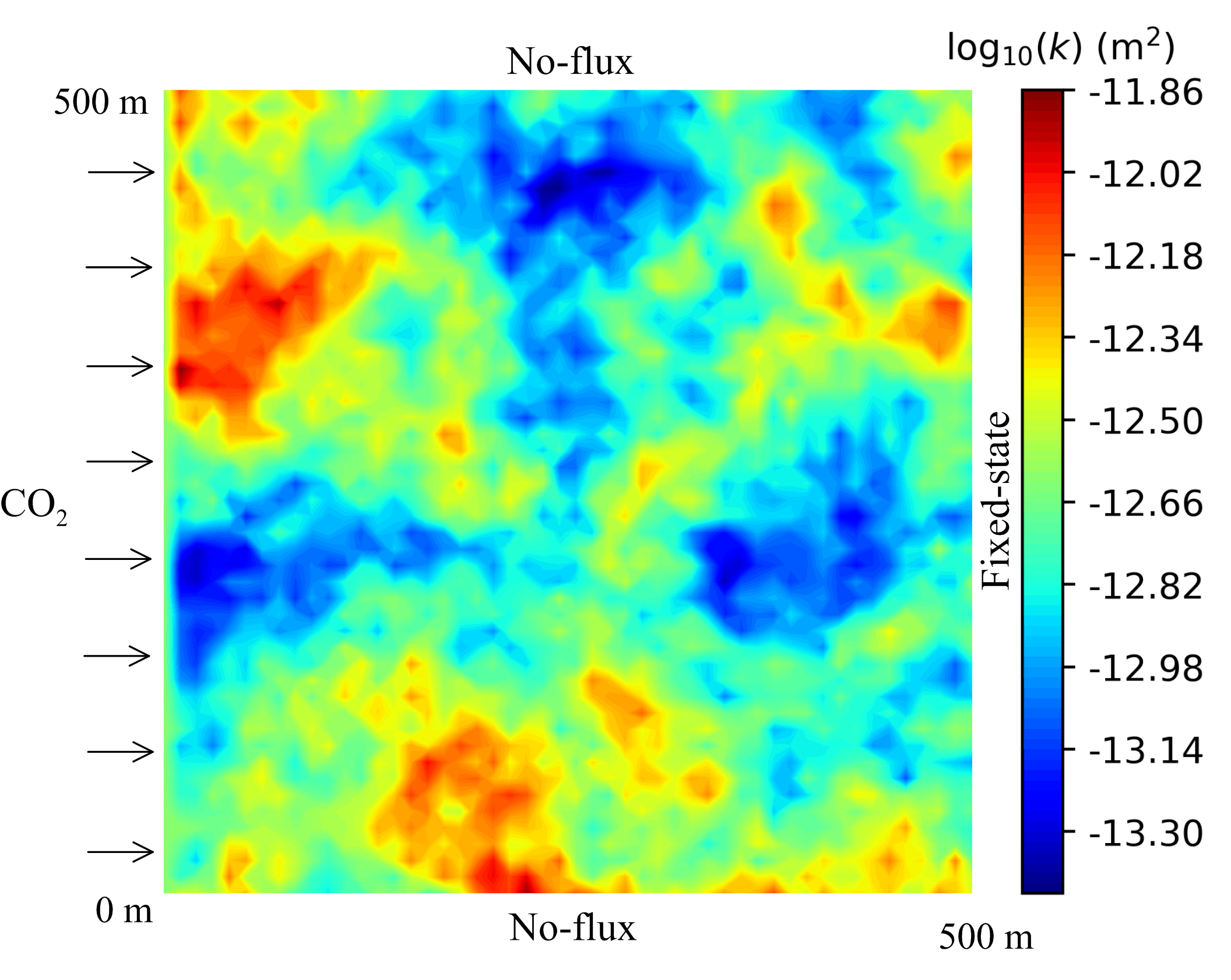}
    \caption{Realization of the permeability field with a correlation length of $\lambda=100$ m (CO$_2$ is injected from the cells on the left boundary).}
\label{Perm}
\end{figure}

To demonstrate the effectiveness and efficiency of our approach, we apply Algorithm~\ref{2stage} to a synthetic study of a geological carbon storage simulation. The synthetic model simulates the migration of CO$_2$ within a 2-D domain. As shown in Figure~\ref{Perm}, the $500$ m$\times$500 m aquifer domain lies in the $x-y$ plane with a grid size of $10$ m and a thickness of $1$ m. The initial pressure is assumed to be $12$ MPa with a constant temperature of $45$ $^\circ$C and a salinity of $15 \%$ by weight. The upper and lower boundaries are assumed to be with no-flux. The CO$_2$ is injected into the 2-D aquifer at a constant rate of $0.35$ kg/s through $50$ injection wells located on the left boundary, while the right boundary is assumed to be fixed-state with a constant hydraulic pressure. The forward simulation is performed using TOUGH2/ECO2N~\citep{pruess2005,pruess1999} and the hydrogeologic properties used are summarized in Table~\ref{ModPar}.
The van Genuchten-Mualem relative permeability function~\citep{corey1954,Mualem1976,van1980} and van Genuchten capillary 
pressure function~\citep{van1980} are used. 
One single simulation with a simulation time of $200$ days takes about $11$ minutes on a Xeon E5-2660 $2.2$ GHz CPU. 

\begin{table}[h!]
\centering
    \caption{Parameters for TOUGH2/ECO2N Numerical Simulations}
\label{ModPar}
\begin{tabular}{lc}
\hline
CO$_2$ injection rate                  & $0.35$ kg/s                                \\
Reference permeability, $k_{\rm{ref}}$ & $2.5 \times 10^{-13}$ $\rm{m}^2$         \\
Porosity (constant)                    & $0.2$                                      \\
Initial pressure                       & $12$ MPa                                  \\
Temperature (isothermal)               & $45$ $^\circ$C                             \\
Salinity                               & $0.15$                                     \\
\textbf{Relative permeability}         &                                          \\
\quad Irreducible water saturation     & $0.2$                                      \\
\quad Irreducible gas saturation       & $0.05$                                     \\
\quad Pore size distribution index     & $0.457$                                    \\
\textbf{Capillary pressure}            &                                          \\
\quad Irreducible water saturation     & $0.0$                                      \\
\quad Pore size distribution index     & $0.457$                                    \\
\quad Air entry pressure               & $1.961\times10^4$ Pa                     \\
\hline
\end{tabular}
\end{table}

The objective of this benchmark problem is to quantify the uncertainty of the evolution of the pressure buildup and CO$_2$ saturation in the $50 \times 50$ cells over time. It is assumed that the uncertainty is only caused by the random heterogeneous permeability field (i.e., $d_x=1$). The application of the model to problems with multiple types of random input fields (i.e., $d_x>1$) is straightforward without requiring any modifications of the network structure. The pressure response $P$ in this example is always larger than the initial pressure, i.e., $1.2\times10^7$ Pa, which is too large compared to the CO$_2$ saturation. This  may lead to an unstable neural network approximation, thus we rescale the pressure values as follows
 \begin{linenomath*}
\begin{equation}
    P'=\frac{P}{10^7}-1.2.
\end{equation}
\end{linenomath*}

In TOUGH2, the permeability heterogeneity is specified by assigning a permeability modifier $\alpha$ to each cell $i$, i.e.,
 \begin{linenomath*}
\begin{equation}
    k_i=\alpha_{i}k_{\rm{ref}},
\end{equation}
\end{linenomath*}
\noindent here $k_{\rm{ref}}=2.5 \times 10^{-13}$ $\rm{m}^2$ is the reference permeability. The log-permeability field is assumed to be a Gaussian random field, thus the permeability modifier
 \begin{linenomath*}
\begin{equation}
    \alpha(\textbf{\emph s})=\exp(G(\textbf{\emph s})),\quad G(\cdot)\sim{N(m,C(\cdot,\cdot))}.
\end{equation}
\end{linenomath*}
\noindent Here, $m=0$ is the constant mean and an exponentiated quadratic covariance function is used, i.e.,
 \begin{linenomath*}
\begin{equation}
    C(\textbf{\emph s},\textbf{\emph{s$'$}})=\sigma_G^2\exp(\frac{-\|{\textbf{\emph s}}-{\textbf{\emph{s$'$}}}\|_2}{\lambda}),
\end{equation}
\end{linenomath*}
\noindent where  $\textbf{\emph{s}}$ and $\textbf{\emph{s$'$}}$ are two arbitrary locations in the domain, and $\sigma_G^2 = 0.5$ and $\lambda = 100$ m are the variance and correlation length, respectively. Note that if this random field is parameterized using the Karhunen-Lo{\` e}ve expansion~\citep{ZHANG2004773}, $\approx 800$ expansion terms will be needed to preserve $\approx 95\%$ of the field variance. In the numerical examples, we directly treat the permeability at each spatial grid as an uncertain variable without using any dimensionality reduction methods, leading to an input dimension of $2500$ (thus all pixel values of the permeability field are taken as uncorrelated). 

One random realization of the permeability field is shown in Figure~\ref{Perm} and snapshots of the corresponding simulated pressure and CO$_2$ saturation at $3$ time instances ($100$, $150$, and $200$ days) after CO$_2$ injection are respectively shown in the first rows of Figure~\ref{P_SgN400_2stage}. It is observed that the pressure evolves relatively smoothly as it is generally insensitive to small-scale variability of the permeability~\citep{Kitanidis2015}. In contrast, the permeability heterogeneity shows a significant influence on the migration of CO$_2$. The saturation displays a complicated spatial variability (strongly nonlinear) and a sharp (discontinuous) front moving at varying speed. This strong nonlinearity and front discontinuity make the development of a surrogate model quite a challenging task.

\subsubsection{Neural Network Design and Performance Evaluation Metrics}
\label{sec:NetDesign}
\begin{figure}[h!]
\centering
\includegraphics[width=0.7\textwidth]{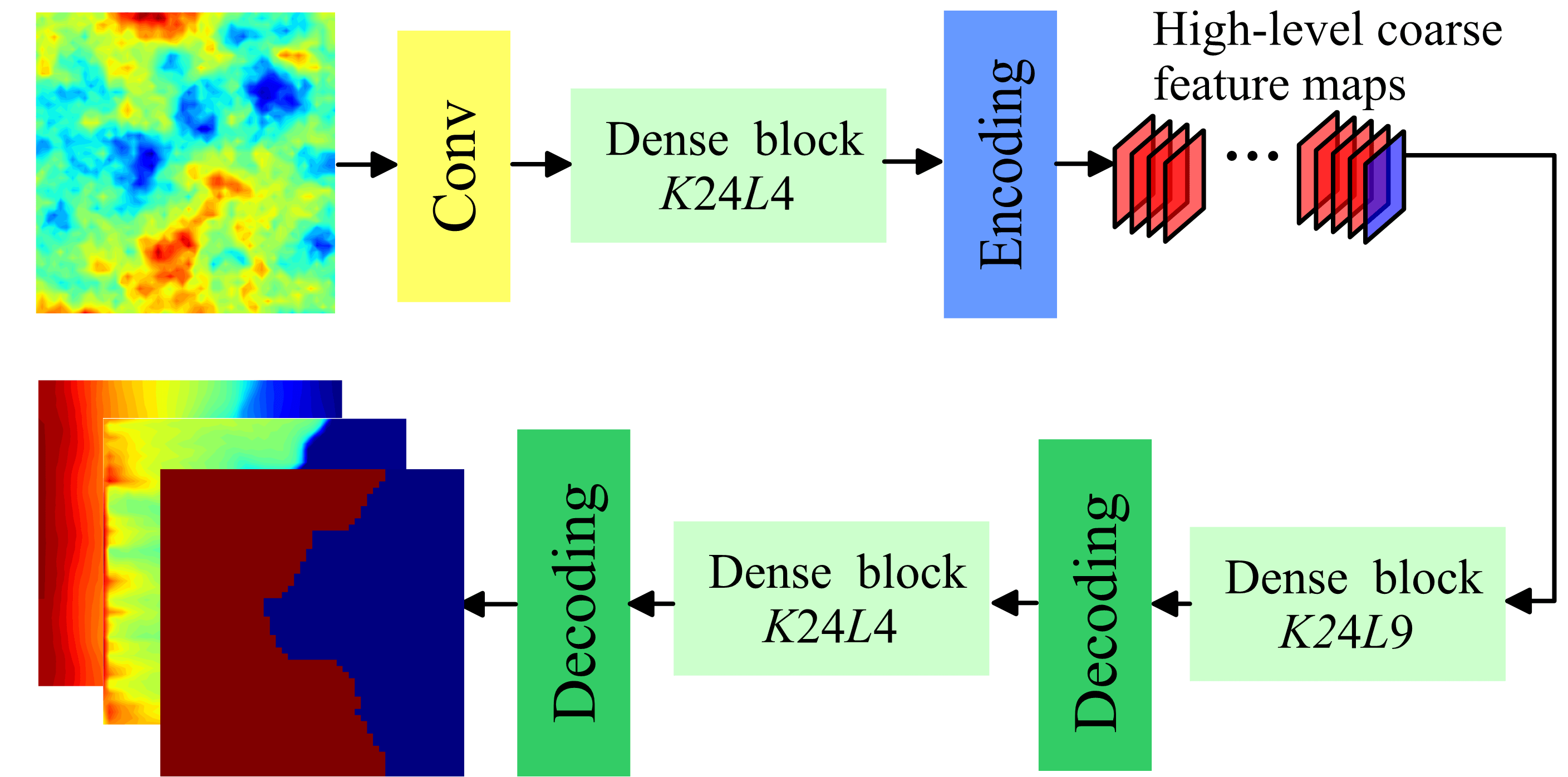}
    \caption{The network configuration used in the numerical examples. It contains three dense blocks with $L=4$, $9$, and $4$ internal layers, and a constant growth rate $K=24$. There are two downsampling layers (Conv and Encoding), thus the dimension of the high-level coarse feature maps is $13\times 13$. The input time $t$ enters the network as an extra feature map (the blue box) of the high-level coarse feature maps.}
\label{DCEDN_used}
\end{figure}

For the design of the network architecture, we follow the guidelines provided in~\citet{zhu2018} and employ the most promising configuration which is illustrated in Figure~\ref{DCEDN_used} with more details on the parameter settings listed in Table~\ref{DCEDN_structure}. The number of initial feature maps after the first convolution layer is $48$. The time $t$ enters the network as an extra feature map of the latent space after the last encoding layer. The network contains three dense blocks with $L=4$, $9$, and $4$ internal layers,  and a constant growth rate $K=24$. The convolution kernels are fixed to be $k'7s2p3$ for the first convolution layer, where the value for each parameter is denoted by the number to the right of the corresponding symbol, $k'3s1p1$ in the convolutional layers within the dense block, $k'1s1p0$ and $k'3s2p1$ for the first and second convolution layers, respectively, in both of the encoding and decoding layers, and $k'4s2p1$ for the transposed convolution layer in the last decoding layer. 

We are interested in the spatiotemporal evolution of the pressure and saturation fields between $100$ and $200$ days after CO$_2$ injection. Thus we collect the output pressure and CO$_2$ saturation at $6$ time instances ($100, 120, 140, 160, 180$, and $200$ days) to train the network. There are $3$ feature maps (i.e., the pressure field and the saturation field together with its corresponding binary representation image) in the output of the last layer (i.e., $[\hat{\mathbf{y}}, \hat{\bm\zeta}] = \mathbf{f}(\mathbf{x}, t) \in\mathbb R^{3\times 50\times 50}$). The sigmoid function is employed as the activation of the output layer to embed the prior knowledge that the saturation and rescaled pressure take values between $0$ and $1$. The initial learning rate $\varepsilon=0.001$, weight decay $\alpha=0.0005$, batch size $M=100$, and binary segmentation  loss weight $w=0.01$. We also use a learning rate scheduler which drops $10$ times on plateau of the root mean squared error. The network is trained for $200$ epochs. 

\begin{table}[]
\centering
    \caption{Network architecture. $N_{\rm{out}}$ denotes the number of output feature maps, and $H_f \times W_f$ denotes the spatial dimension of the feature map.}
\label{DCEDN_structure}
\begin{tabular}{lcc}
\hline
Layers                  & $N_{\rm{out}}$ & Resolution $H_f\times W_f$ \\
\hline                                                       
Input                   & 1     & $50 \times 50$             \\
Convolution $(k'7s2p3)$ & 48    & $25 \times 25$             \\
Dense Block 1 $(K24L4)$ & 144   & $25 \times 25$             \\
Encoding Layer          & 72    & $13 \times 13$             \\
Dense Block 2 $(K24L9)$ & 289   & $13 \times 13$             \\
Decoding Layer 1        & 144   & $25 \times 25$             \\
Dense Block 3 $(K24L4)$ & 240   & $25 \times 25$             \\
Decoding Layer 2        & 3     & $50 \times 50$             \\
\hline                         
\end{tabular}
\end{table}

To evaluate the quality of the trained surrogate model, we consider two commonly used metrics, i.e., the coefficient of determination ($R^2$) and root mean squared error (RMSE). The $R^2$ metric is defined as
\begin{linenomath*}
\begin{equation}
    R^2=1-\frac{\sum_{i=1}^{N}||\textbf y^i-\hat{\textbf y}^i||_2^2}{\sum_{i=1}^{N}||\textbf y^i-\bar{\textbf y}||_2^2},
\end{equation}
\end{linenomath*}
\noindent where $N$ is the number of samples, $\textbf y^i$ and $\hat{\textbf y}^i$ are the TOUGH2 and deep network predicted outputs, respectively, and $\bar{\textbf y}=1/N\sum_{i=1}^{N}\textbf y^i$. The RMSE metric is written as
\begin{linenomath*}
\begin{equation}
    {\rm RMSE}=\sqrt{\frac{1}{N}\sum_{i=1}^{N}||\textbf y^i-\hat{\textbf y}^i||_2^2}.
\end{equation}
\end{linenomath*}
The $R^2$ metric is a normalized error enabling the comparison between different datasets, with the score closer to $1.0$ corresponding to better surrogate quality, while RMSE is a metric commonly used for monitoring the convergence of both the training and test errors during the training process. 

\subsection{Results}\label{section:R&D}

In this section, we  assess the performance of our model in approximating the time-dependent multi-output of the geological carbon storage multiphase flow system in the case of high-dimensionality and response discontinuity. The approximation accuracy of the model is evaluated using $500$ randomly selected permeability test samples. To evaluate the significance of using the MSE-BCE loss in training our model, we also compare our results with those obtained from a network with the same architecture and training data but using only the MSE loss. Such a network has   only two output images (the pressure and CO$_2$ saturation fields). To distinguish between this network and our proposed network, we refer to them as the network with MSE loss and the network with MSE-BCE loss, respectively. The results obtained in uncertainty quantification tasks using our deep network based model are compared to those computed with vanilla MC.

\subsubsection{Approximation Accuracy Assessment}\label{AAA}

\begin{figure}[h!]
\centering
\includegraphics[width=0.9\textwidth]{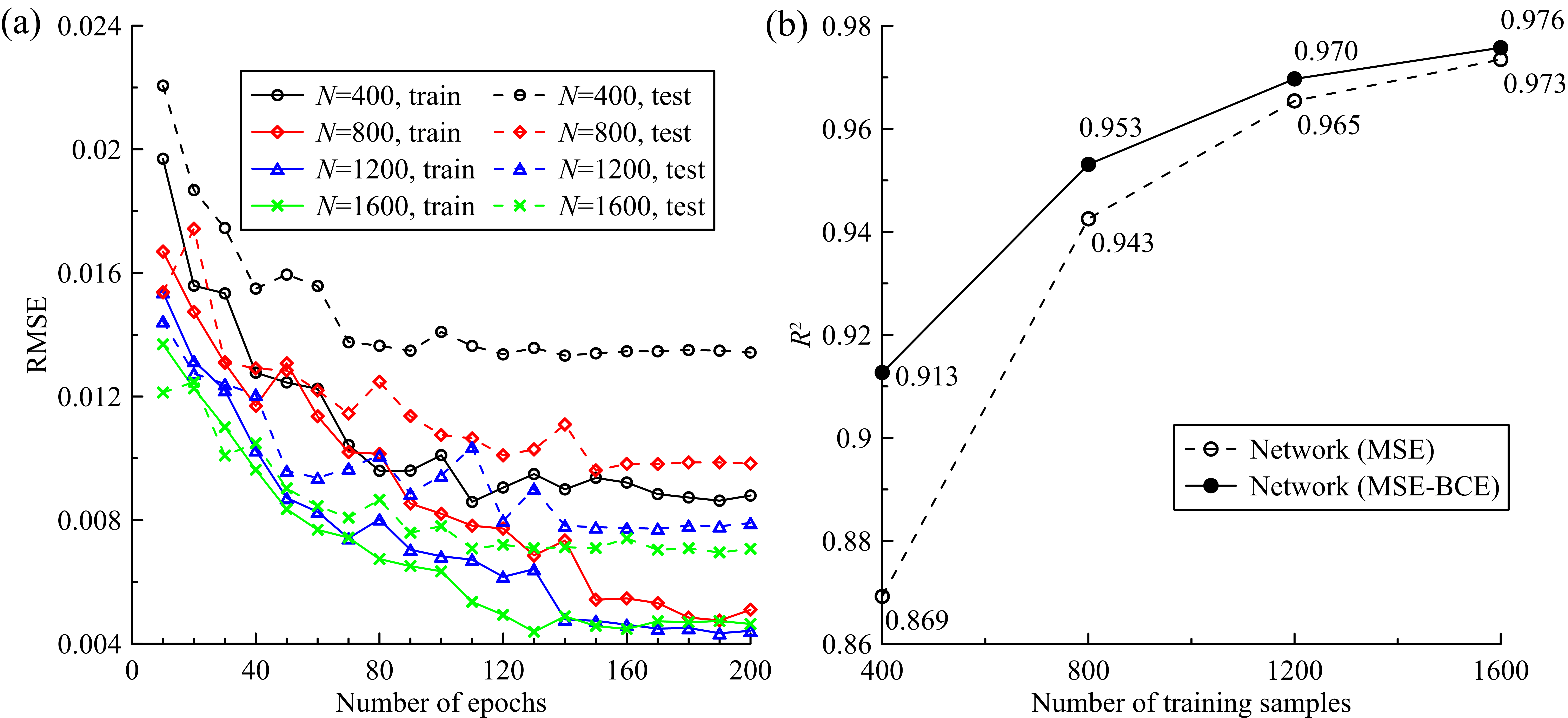}
    \caption{(a) RMSE decay with the number of epochs when training the network with MSE-BCE loss using different training sample sizes. (b) Comparison of the $R^2$ scores of networks trained with MSE and MSE-BCE losses evaluated on $500$ test samples with the number of training samples.}
\label{RMSE_R2}
\end{figure}

To illustrate the convergence of the approximation error  with respect to the training sample size, we generate four training sample sets with $400$, $800$, $1200$, and $1600$ model evaluations. Figure~\ref{RMSE_R2}a plots the RMSE decay with the number of epochs during the training process. The model is trained on a NVIDIA GeForce GTX $1080$ Ti X GPU which requires about $900-2500$ seconds for training $200$ epochs when the training sample size varies from $400$ to $1600$. Each sample includes system responses at $n_t=6$ time instances as discussed in section~\ref{sec:NetDesign}. It is observed that the RMSE starts to stabilize after $150$ epochs in all cases. The $R^2$ score for the test dataset for each surrogate with different training dataset sizes is shown in Figure~\ref{RMSE_R2}b. The figure shows that with only $400$ training samples the model achieves a relatively high $R^2$ value of $0.913$ for a problem with $2500$ stochastic input dimensions and in the presence of response discontinuity. When increasing the samples size to $1600$, the model  achieves a $R^2$ value of $0.976$. 

Figure~\ref{RMSE_R2}b clearly demonstrates that the model using the MSE-BCE loss achieves a higher $R^2$ value for a given dataset size than the network trained with the MSE loss. 
 With  $400$ training samples, the network with MSE loss only achieves a $R^2$ value of $0.869$, much lower than that of the MSE-BCE loss based network ($0.913$). As more training samples are available, the difference between the $R^2$ scores of the networks for the two loss functions decreases. The results indicate that the MSE-BCE loss training strategy substantially improves the model's performance in approximating the multiphase flow GCS model especially when the training sample size is small.  

\begin{figure}[h!]
	\centering
	\includegraphics[width=0.95\textwidth]{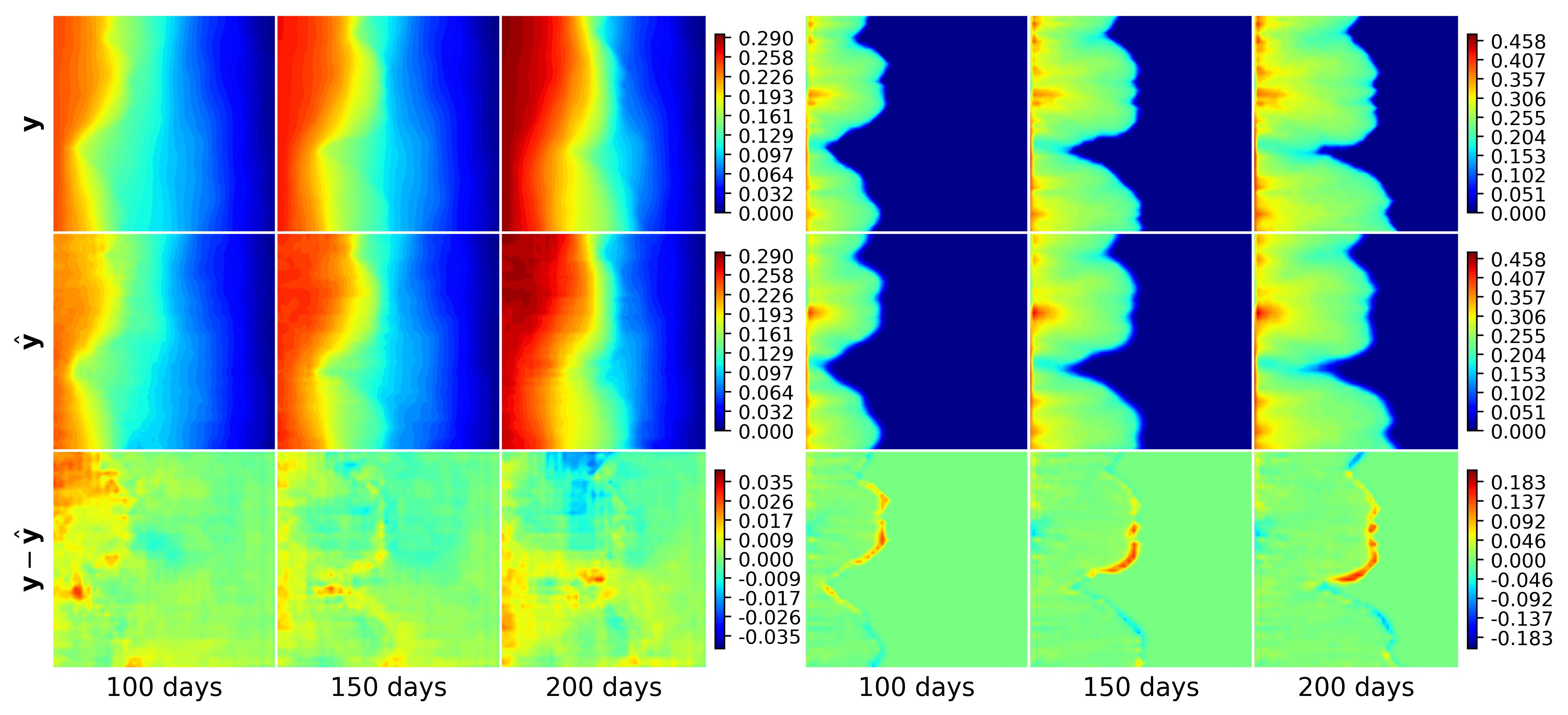}
	\caption{Snapshots of the pressure (left) and CO$_2$ saturation (right) fields at $100$, $150$, and $200$ days predicted by TOUGH2 ($\mathbf{y}$) and the network with MSE-BCE loss ($\hat{\mathbf{y}}$) for the testing permeability realization shown in Figure~\ref{Perm}. We run TOUGH2 for $400$ training permeabilty samples and use the corresponding outputs at $6$ time instances ($100, 120,\ldots, 200$ days) to train the network. Note the accuracy of the model prediction at $150$ days when no training data are provided.}
	\label{P_SgN400_2stage}
\end{figure}

\begin{figure}[h!]
	\centering
	\includegraphics[width=0.95\textwidth]{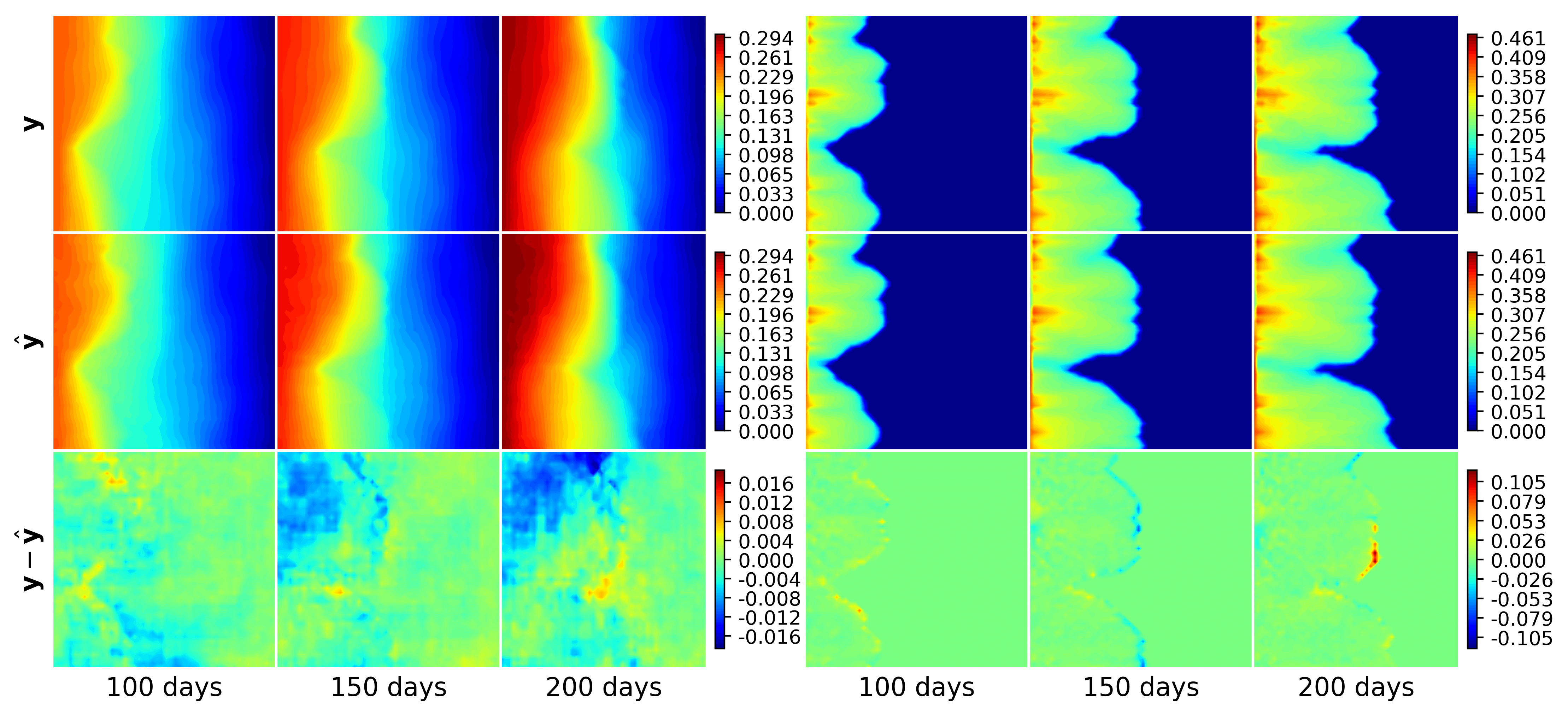}
	\caption{Snapshots of the pressure (left) and CO$_2$ saturation (right) fields at $100$, $150$, and $200$ days predicted by TOUGH2 ($\mathbf{y}$) and the network with MSE-BCE loss ($\hat{\mathbf{y}}$) for the permeability field shown in Figure~\ref{Perm} using $1600$ training samples with model outputs at $6$ time instances ($100, 120,\ldots, 200$ days). The model outputs at $150$ days are not considered when training the network.}
	\label{P_SgN1600_2stage}
\end{figure}

The performance of our model in accurately approximating the multiphase flow model is further illustrated in Figures~\ref{P_SgN400_2stage} and~\ref{P_SgN1600_2stage}, which depict a comparison of the pressure and CO$_2$ saturation fields at $100, 150$, and $200$ days predicted by TOUGH2 and our network model using $400$ and $1600$ training samples, respectively. These predictions refer to the permeability realization shown in Figure~\ref{Perm} that is  randomly selected from the test set. In both cases, the model as expected achieves higher approximation accuracy for the relatively smooth pressure field than  for the strongly nonlinear and discontinuous saturation field. Even when using only $400$ training samples, the model is capable of capturing the nonlinear pressure field. It also  reproduces the front discontinuity at all three time instances although with a higher approximation error. These are well-recognized challenges for other surrogate models~\citep{LiaoZhang2013,LiaoZhang2014,Xiu2005}. The relatively large approximation error of the model when using $400$ training samples is mainly caused by the large error in the approximation of the saturation front. When increasing the training sample size to $1600$, as shown in Figure~\ref{P_SgN1600_2stage}, the model provides better characterization of the local nonlinear features as well as of the position of the saturation front. 

\begin{figure}[h!]
\centering
	\includegraphics[width=0.95\textwidth]{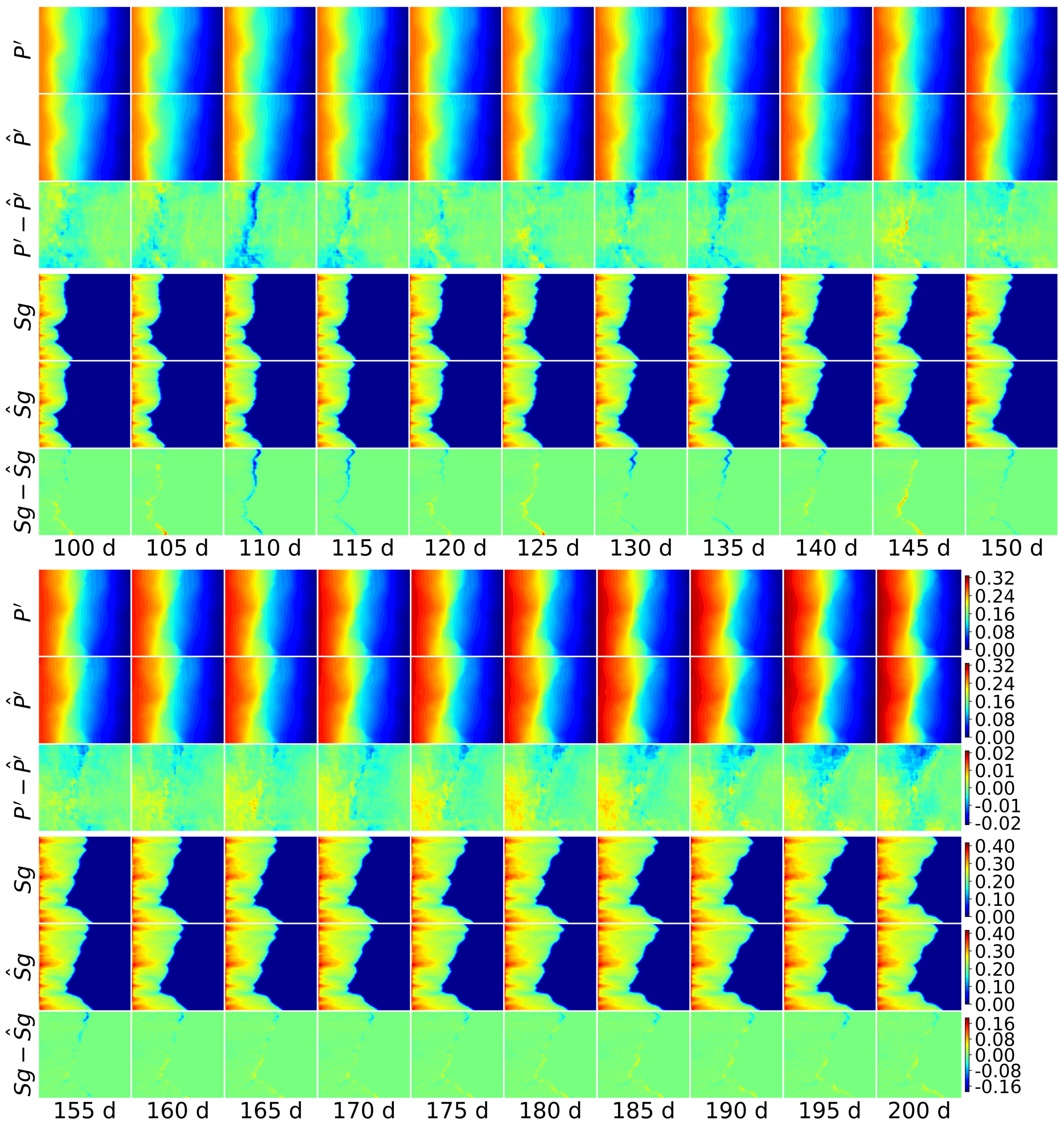}
	\caption{Computed and reference time-dependent outputs at several time instances. The model is trained with MSE-BCE loss using $1600$ samples with the output pressure ($P'$) and CO$_2$ saturation ($Sg$) at only $6$ time instances ($100, 120, 140, 160, 180, 200$ days). The corresponding permeability field (not shown) is randomly selected from the test dataset. $\hat{P}'$ and $\hat{S}g$ denote the predictions of the network.}
	\label{interp_t6_21}
\end{figure}

Note that our model was trained with output data at $6$ time instances including $100$ and $200$ days but not $150$ days. It is clear that visually there is no significant difference between the approximation errors at time instances when training data were provided and other interpolated time instances in particular $t=150$ days. The model capability to accurately predict the time evolution of the output fields is further illustrated in Figure~\ref{interp_t6_21} which depicts the predicted pressure and CO$_2$ saturation fields at $21$ time instances ranging every $5$ days from $100$ to $200$ days. We can see that the model provides good approximation at all time instances. The prediction errors at the time instances when training data were provided ($100, 120, 140, 160, 180, 200$ days) are similar to those in the remaining predictions at $15$ time instances. Thus by treating the time as an input and training with output data at only a limited number of time instances, the model is capable of approximating the time-dependent multi-output of the dynamic system efficiently. 

\begin{figure}[h!]
	\centering
	\includegraphics[width=0.95\textwidth]{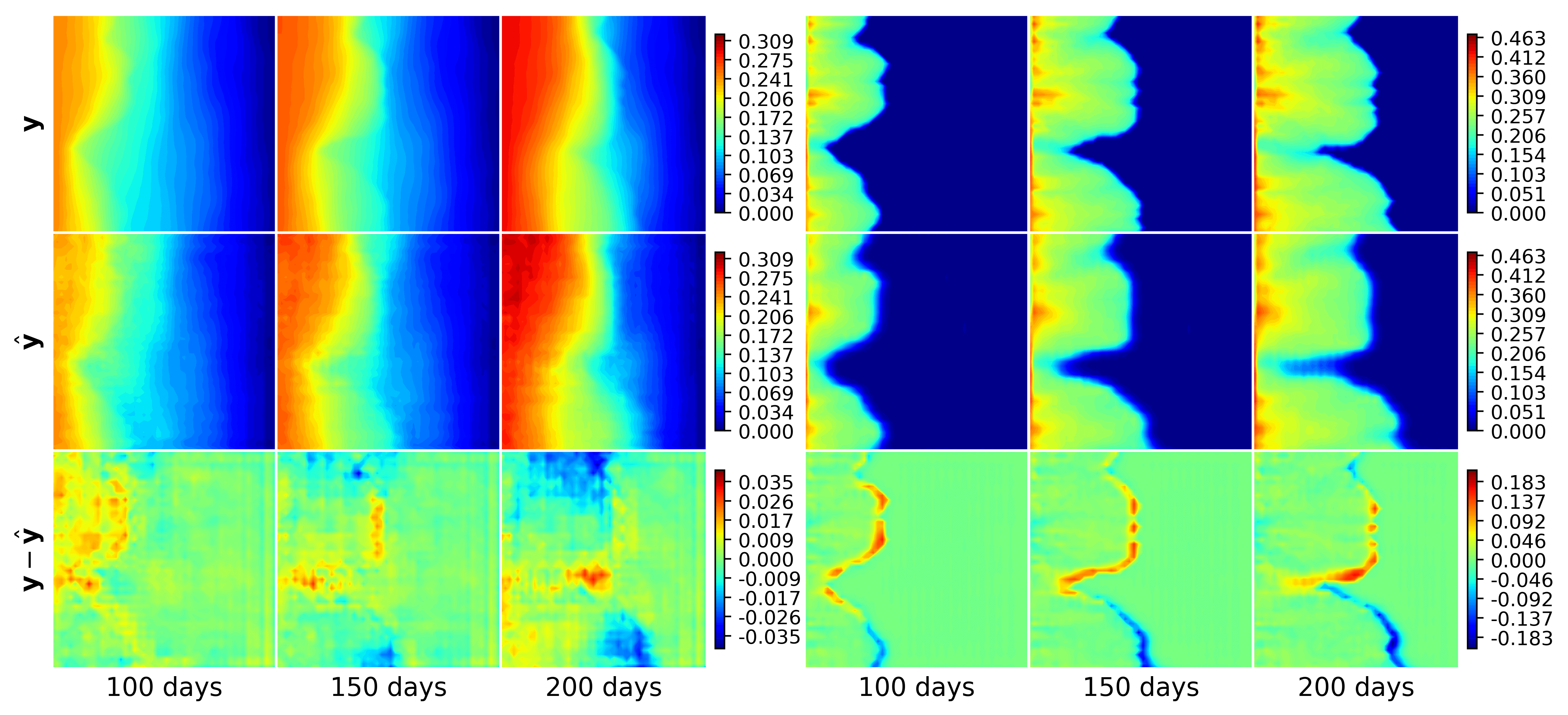}
	\caption{Snapshots of the pressure (left) and CO$_2$ saturation (right) fields at $100$, $150$, and $200$ days predicted by TOUGH2 ($\mathbf{y}$) and the network with MSE loss ($\hat{\mathbf{y}}$) for the permeability field shown in Figure~\ref{Perm} using $400$ training samples with model outputs at $6$ time instances ($100, 120,\ldots, 200$ days). The model outputs at $150$ days are not considered when training the network.}
	\label{P_SgN400_1stage}
\end{figure}

\begin{figure}[h!]
	\centering
	\includegraphics[width=0.95\textwidth]{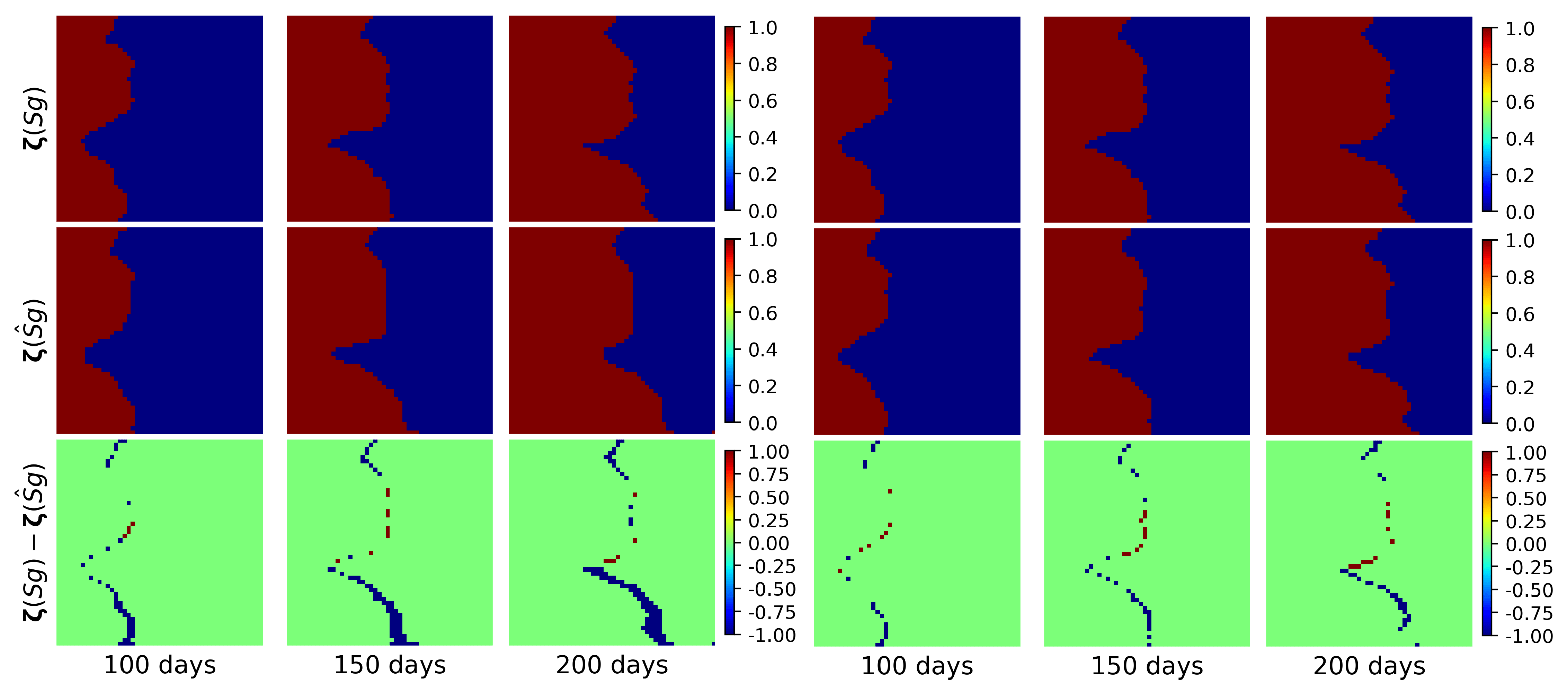}
	\caption{Binary images showing the distribution of the CO$_2$ saturation plumes obtained from binarizing the saturation fields in Figures~\ref{P_SgN400_1stage} (on the left, using MSE loss) and~\ref{P_SgN400_2stage} (on the right, using MSE-BCE loss) using equation~(\ref{Closs}). $Sg$ and $\hat{S}g$ denote the CO$_2$ saturation predicted by TOUGH2 and the network, respectively.}
	\label{label_400}
\end{figure}

Figure~\ref{P_SgN400_1stage} shows the predictions of the MSE loss based model with $400$ training samples for the same output fields shown in Figure~\ref{P_SgN400_2stage}. Comparing with Figure~\ref{P_SgN400_2stage}, it can be seen that visually there is no significant difference in the  pressure and saturation fields approximation errors between the two models except near the saturation front. However, it is observed that the MSE-BCE loss based model provides better characterizations for the saturation front positions than the MSE loss based model. This is demonstrated in Figure~\ref{label_400} which compares the binary images obtained from binarizing the saturation fields in Figures~\ref{P_SgN400_2stage} and~\ref{P_SgN400_1stage} using equation~(\ref{Closs}). The pixels with value $1$ reflect the distribution of CO$_2$ plume and the edges between pixels with values $1$ and $0$ correspond to the positions of the saturation front. It is observed that the model with the MSE-BCE loss function provides more accurate prediction for the positions of the  saturation front. For surrogate modeling of the saturation field, underestimating (overestimating) the position of the discontinuous saturation front will lead to an underestimation (overestimation) of the saturation in the regions between the predicted and actual fronts, resulting in large approximation errors.  

\begin{figure}[h!]
	\centering
	\includegraphics[width=0.6\textwidth]{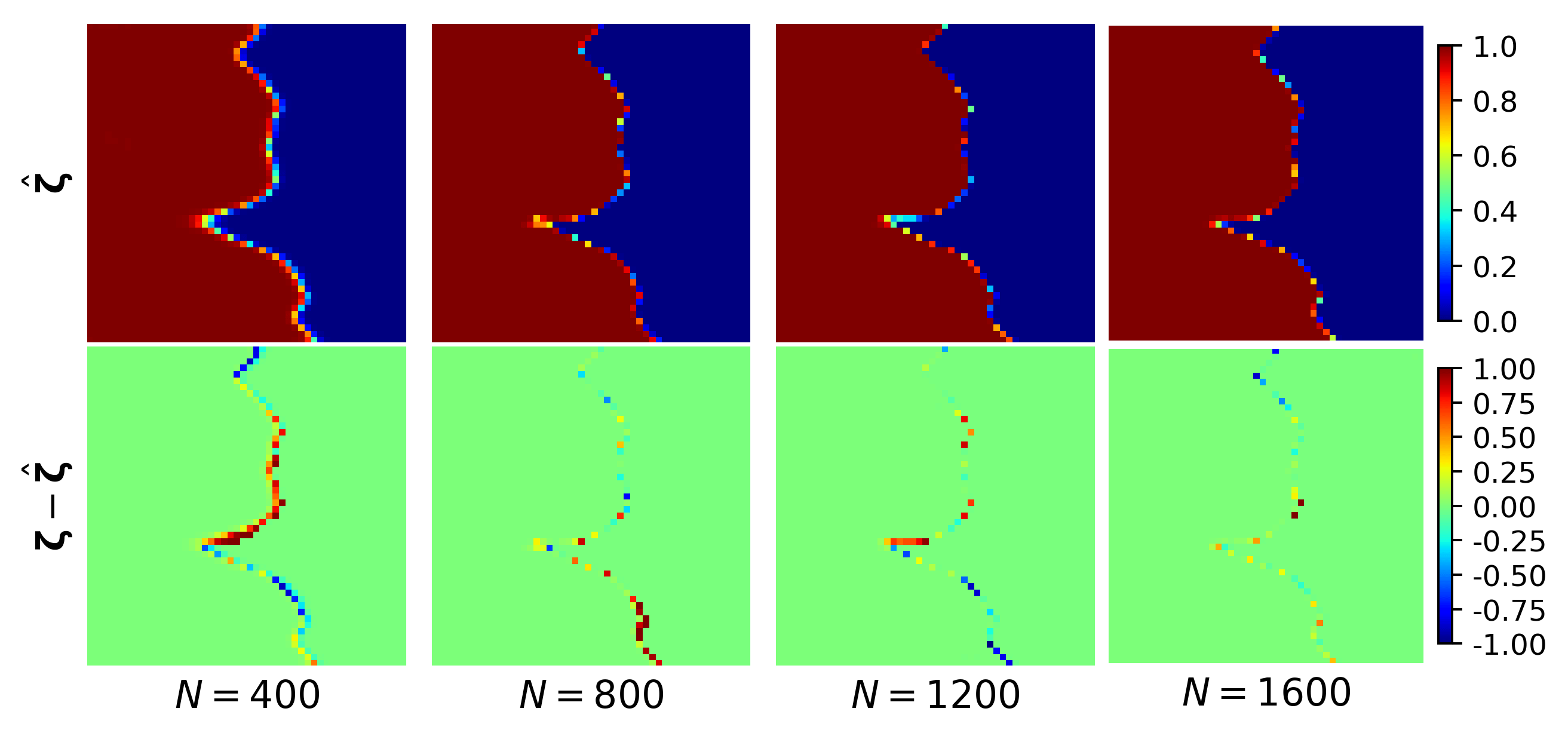}
	\caption{Predictions of the network with MSE-BCE loss trained using four datasets for the binary image at $t=200$ days shown in the upper right of Figure~\ref{label_400}. $(\boldsymbol{\zeta}-\hat{\boldsymbol{\zeta}})$ denotes the difference between the predictions of TOUGH2 and the network.}
	\label{label_err}
\end{figure}

Figure~\ref{label_err} shows the predictions of the MSE-BCE loss function deep network model trained with four different datasets for the binary image at $t=200$ days shown in the upper right of Figure~\ref{label_400}. For the four cases, the approximations for the binary image are rather accurate except near the saturation front. The BCE loss is mostly induced by the approximation errors in regions around the saturation front and considering it in addition to the standard MSE loss promotes a more accurate approximation of the discontinuous front. 

\begin{figure}[h!]
	\centering
	\includegraphics[width=0.5\textwidth]{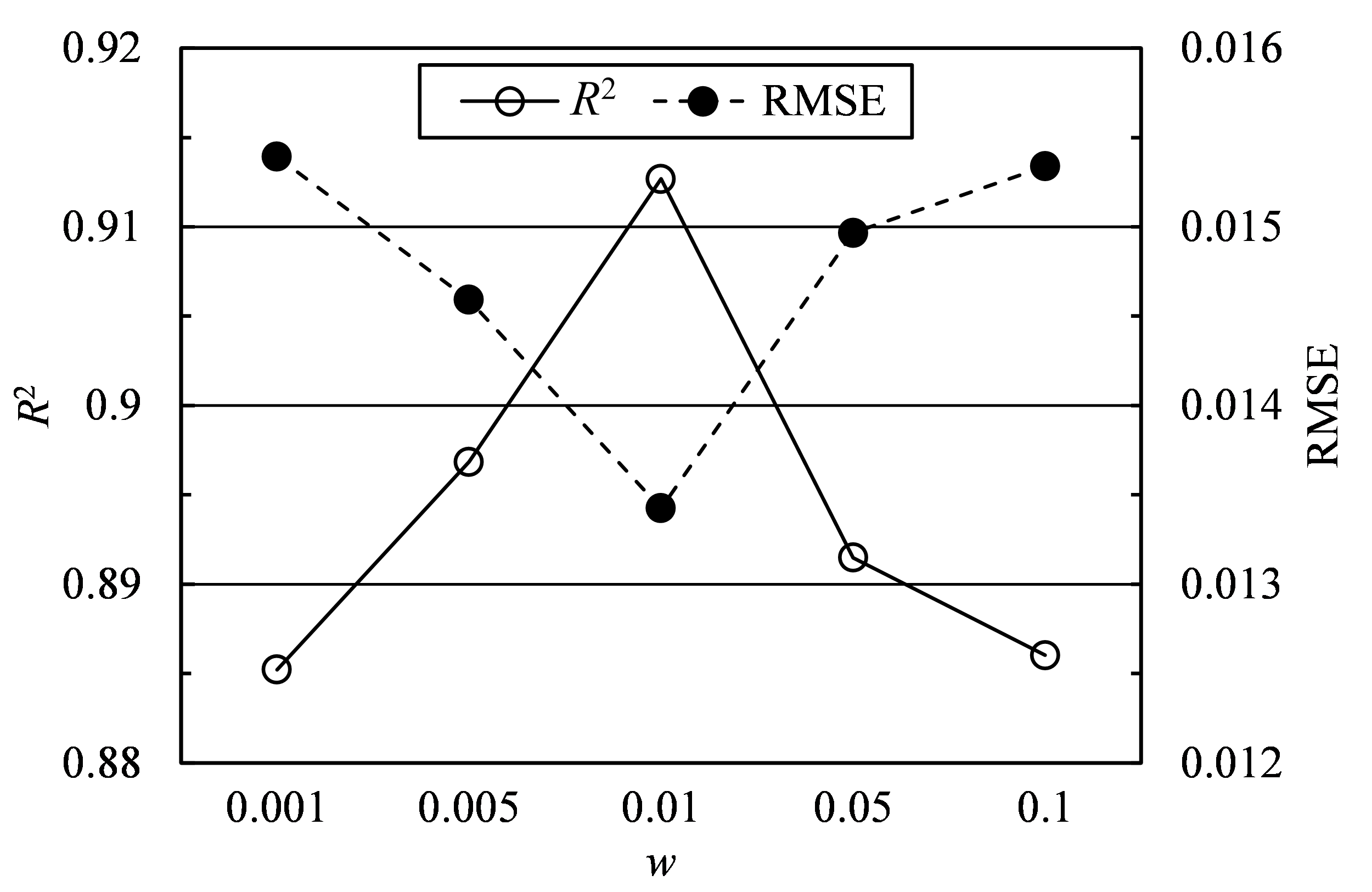}
	\caption{$R^2$ scores and RMSEs of the network with MSE-BCE loss evaluated on $500$ test samples with different BCE loss weights  $w$. The networks are trained using $400$ training samples.}
	\label{w_r2_rmse}
\end{figure}

\begin{figure}[h!]
	\centering
	\includegraphics[width=0.5\textwidth]{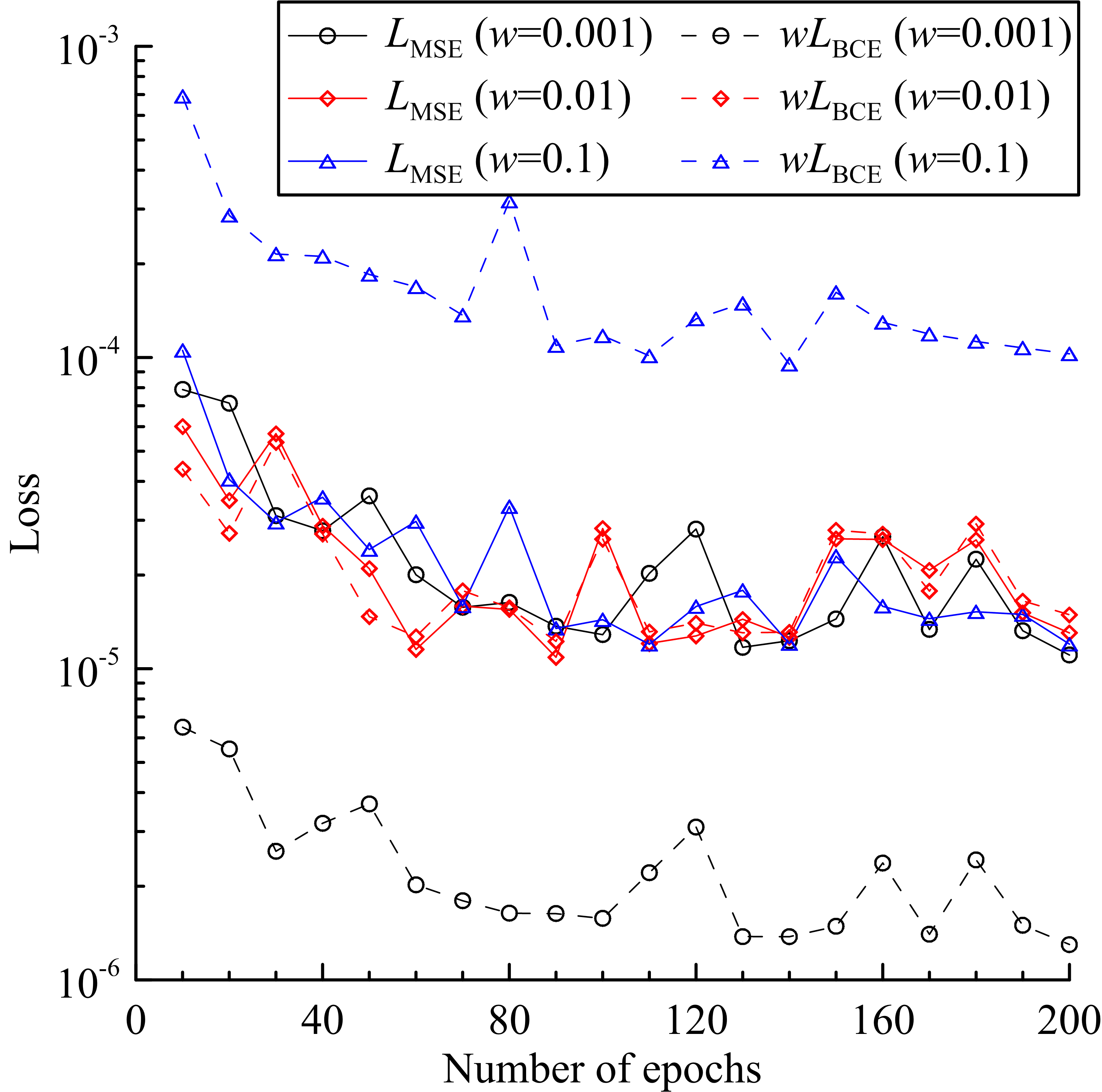}
	\caption{MSE and weighted BCE losses decay with the number of epochs when training the network with different BCE loss weight $w$ using $400$ training samples.}
	\label{MSE-BCE}
\end{figure}

In the above experiments, the BCE loss weight $w$ was chosen as $0.01$. This weight decides the influence of the BCE loss on the hybrid loss in equation~(\ref{eq:Hybridloss}). It is understandable that a lager $w$ will promote a better approximation of the discontinuous saturation front. However, our objective in surrogate modeling of multiphase flow models is not only to accurately characterize the discontinuous saturation front, but more importantly to provide globally accurate approximation for the entire output fields. The value of $w$ should be chosen carefully. To investigate the influence of $w$ on the performance of our method, we test five different values of $w$ ranging from $0.001$ to $0.1$. The corresponding results are shown in Figure~\ref{w_r2_rmse}, which depicts the $R^2$ scores and RMSEs of networks with different $w$ values evaluated on $500$ test samples. It is observed that among the five surrogates the one with $w=0.01$ performs best, and that the performance decreases when increasing or reducing the weight. We further analyze the decay of the values of the MSE loss and weighted BCE loss during the training process. As shown in Figure~\ref{MSE-BCE}, the network with $w=0.01$ has similar MSE and weighted BEC loss values during the training process, thus putting similar weights on global regression of the output fields and approximation of the saturation front. In contrast, the networks with $w=0.001$ and $0.1$ focus more on global regression (larger MSE loss than weighted BCE loss) and local refinement (larger weighted BCE loss than MSE loss), respectively. According to these results, a value $w$ is recommended that equally weighs the MSE and BCE losses.

Recall that we use a dense fully convolutional encoder-decoder network architecture in our method to transform the surrogate modeling task to an image-to-image regression problem. The above results indicate that, first of all, the method shares the good property of CNNs in image processing, making it robust in handling high-dimensional input and output fields (images). Secondly, the use of deep neural networks provides high flexibility and robust capability for surrogate modeling of strongly nonlinear and discontinuous responses. In addition, it was shown that treating time as an additional input to the network, the model can efficiently provide prediction of time-dependent outputs of dynamic systems at any time instance. Finally, the densely connected convolutional network architecture (i.e., dense block) was shown to substantially enhance the exploitation of local information of data and improve the feature flow through the network, which has also been illustrated in~\citet{huang2017} and~\citet{zhu2018}. As a result, the developed model was able to provide relatively good approximation with limited training samples effectively resolving the three challenges associated with surrogate modeling of multi-phase flow models discussed in section~\ref{intro}.

\begin{figure}[h!]
\centering
\includegraphics[width=0.95\textwidth]{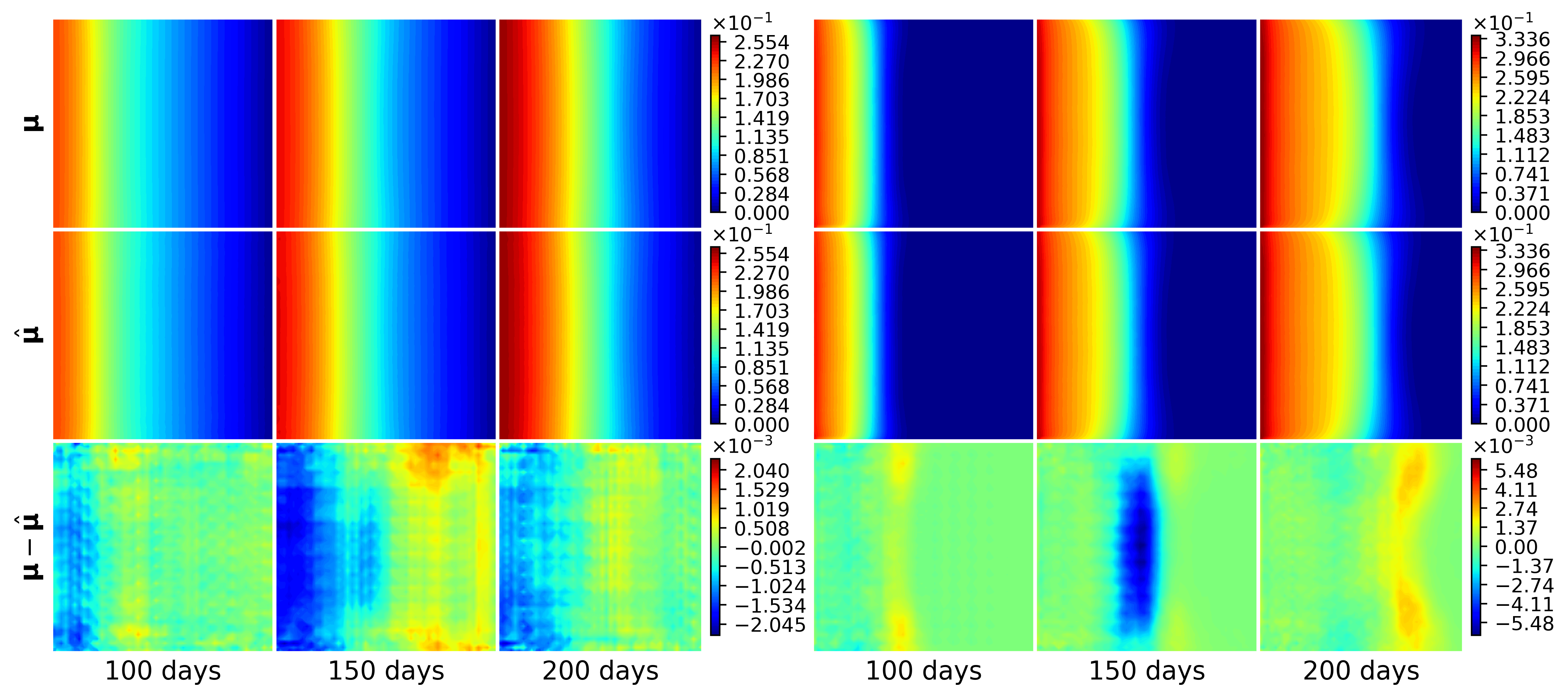}
    \caption{Mean fields of pressure (left) and CO$_2$ saturation (right) obtained by MC sampling ($\boldsymbol{\mu}$) and the network with MSE-BCE loss ($\hat{\boldsymbol{\mu}}$) at $3$ time instances ($100, 150, 200$ days) after CO$_2$ injection. The numbers of model runs needed by MC and for the training of network are $20000$ and $1600$, respectively.}
\label{mean}
\end{figure}

\subsubsection{Uncertainty modeling}

In this section, we test our model's effectiveness and efficiency in addressing UQ tasks for the  GCS model. The MSE-BCE loss based network model is trained with $1600$ samples. Uncertainty quantification is performed using MC approximations. Monte Carlo sampling is conducted on the cheap-to-evaluate  surrogate instead of the computationally intensive GCS model. This means that the deep network is used as a full replacement of the GCS model and thus no extra model evaluations are required. The computational cost is mainly from the evaluation of the $1600$ training samples needed for the surrogate construction. We randomly generate $20000$ MC realizations by direct GCS model runs to compute the reference statistics. We use the deep network and TOUGH2 to predict the system response for the same input realizations when performing UQ. In this way, the possible inaccuracy in the UQ solution will only be caused by the approximation error of the surrogate model.

\begin{figure}[h!]
\centering
\includegraphics[width=0.95\textwidth]{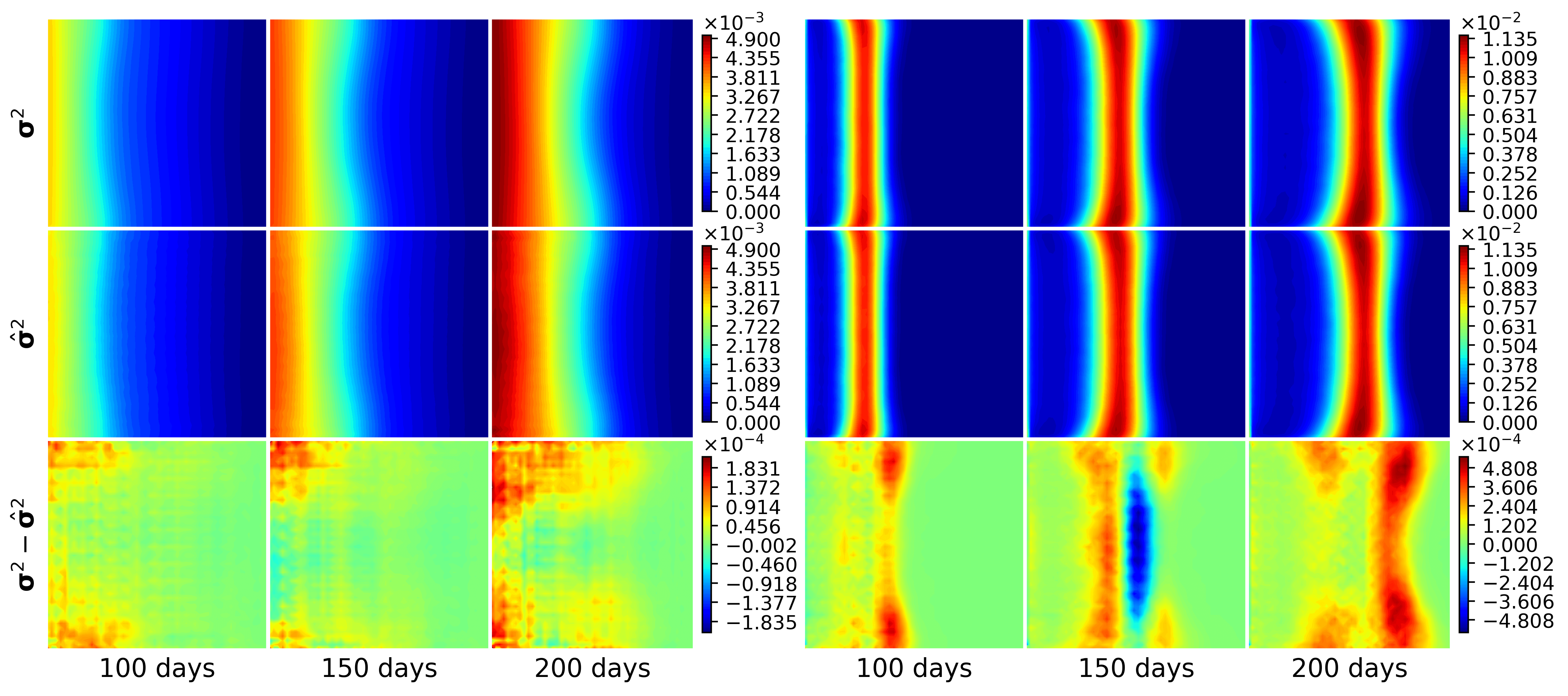}
    \caption{Variance of pressure (left) and CO$_2$ saturation (right) obtained by MC sampling ($\boldsymbol{\sigma}^2$) and the network with MSE-BCE loss (($\hat{\boldsymbol{\sigma}}^2$)) at $3$ time instances ($100, 150, 200$ days) after CO$_2$ injection. The numbers of model runs needed by MC and for the training of network are $20000$ and $1600$, respectively.}
\label{variance}
\end{figure}

Figures~\ref{mean} and~\ref{variance} illustrate the mean and variance of the pressure and CO$_2$ saturation fields at $100, 150$, and $200$ days, respectively. As CO$_2$ is injected from the left boundary with an equal injection rate for all cells, the isolines of mean fields are approximately parallel to the $y$-axis (Figure~\ref{mean}) and as expected much smoother than those corresponding to a single permeability realization. However, the variance of the CO$_2$ saturation around the front is relatively large with large gradients because of the presence of the discontinuity. It can be seen that both the mean and variance fields predicted by the  model including those at  time instances (e.g., $150$ days) where no training data were provided are very close to the reference solutions.

\begin{figure}[h!]
\centering
\includegraphics[width=0.6\textwidth]{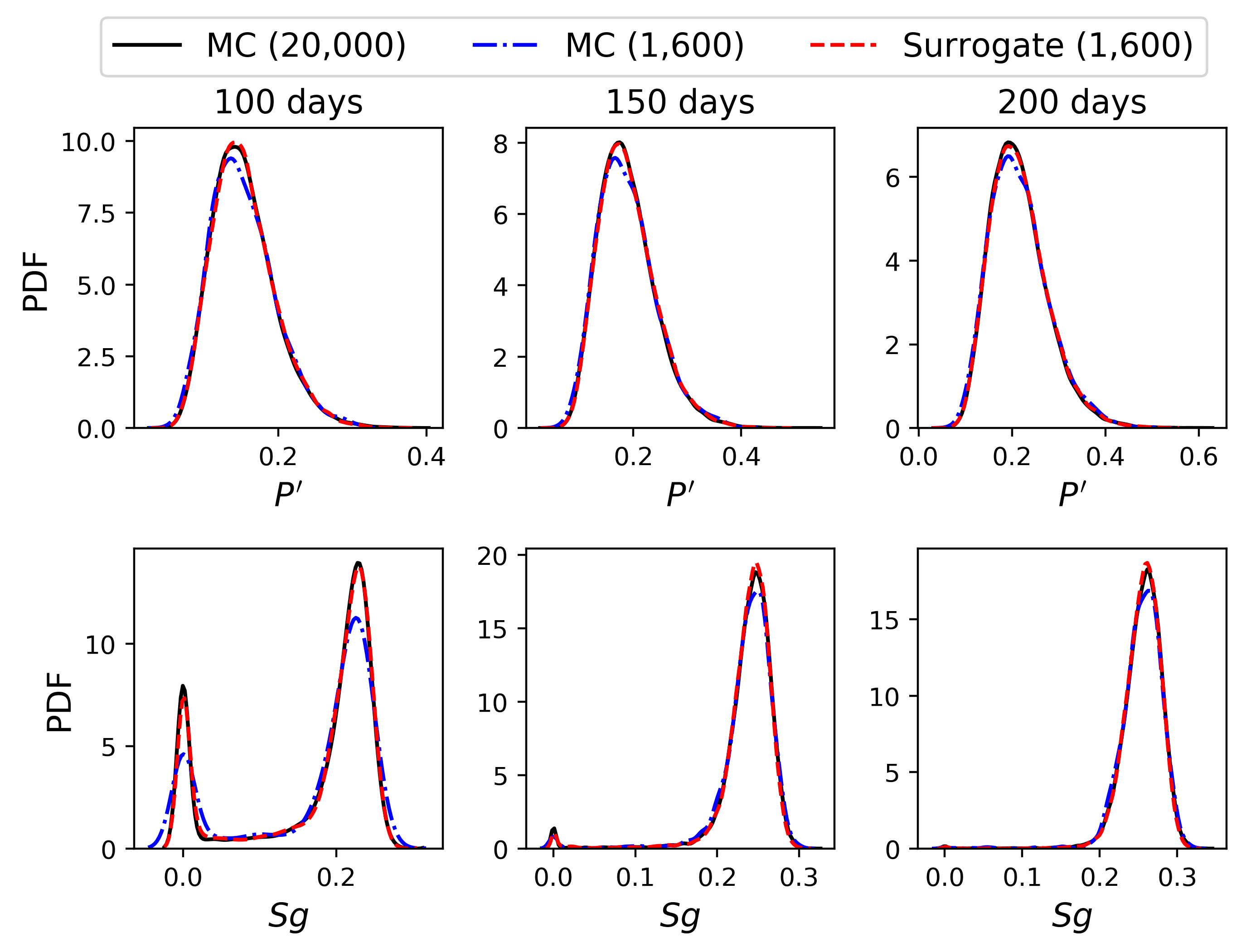}
    \caption{PDFs of pressure ($P'$) and CO$_2$ saturation ($Sg$) at location ($120$ m, $250$ m) at $100, 150$, and $200$ days. The numbers of model runs needed by MC and the surrogate method are shown in parenthesis.}
\label{PDF_p_Sg1}
\end{figure}

\begin{figure}[h!]
\centering
\includegraphics[width=0.6\textwidth]{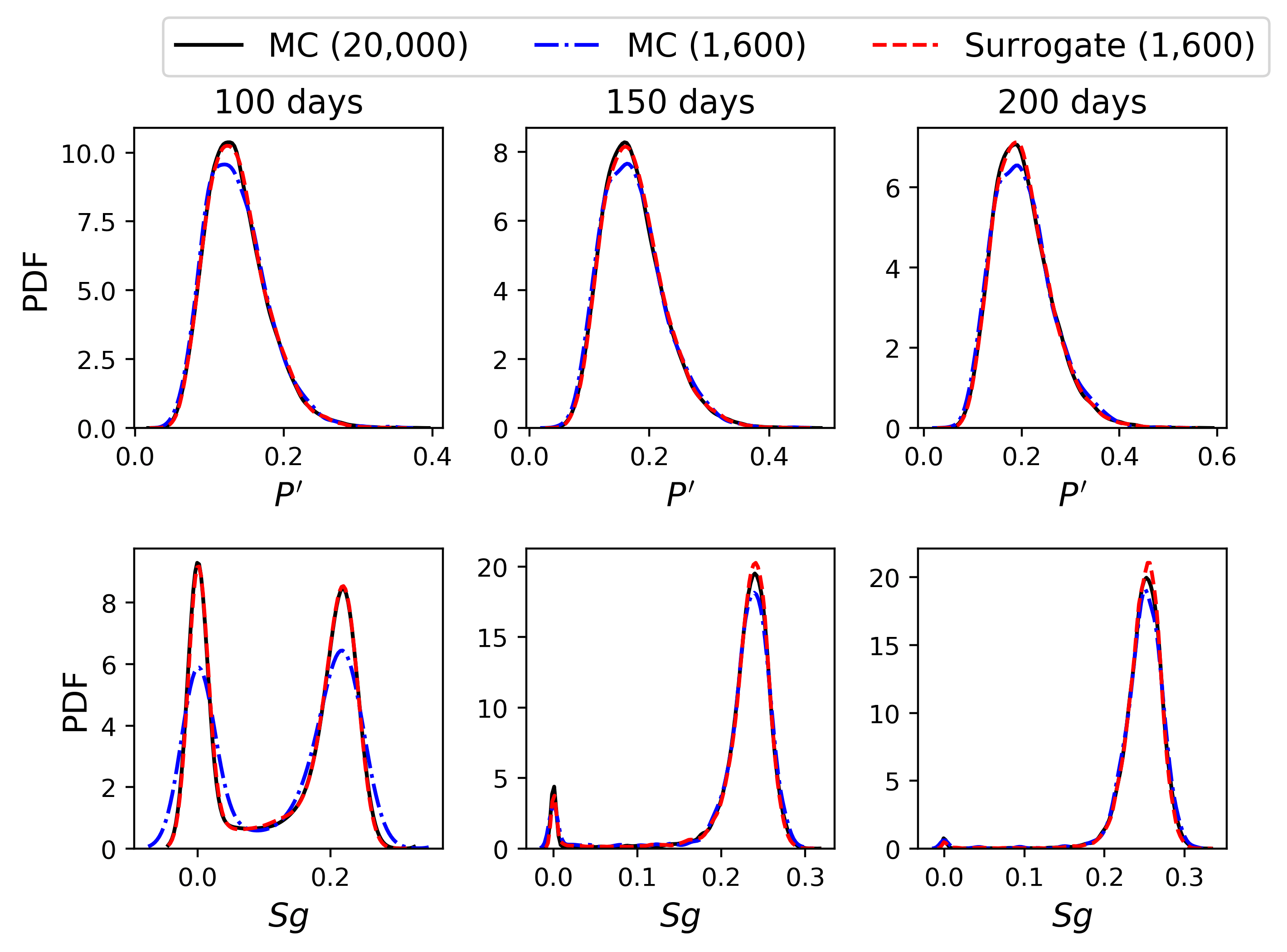}
    \caption{PDFs of pressure ($P'$) and CO$_2$ saturation ($Sg$) at location ($140$ m, $100$ m) at $100, 150$, and $200$ days. The numbers of model runs needed by MC and the surrogate method are shown in parenthesis.}
\label{PDF_p_Sg2}
\end{figure}

Next, we look at the PDFs of the pressure and CO$_2$ saturation. Figures~\ref{PDF_p_Sg1} and~\ref{PDF_p_Sg2} show the PDFs of pressure and CO$_2$ saturation at $3$ times instances at locations ($120$ m, $250$ m) and ($140$ m, $100$ m), respectively. We also plot the estimated PDFs from MC sampling using the same number (i.e., $1600$) of model executions as we used in the network training. This would allow us to compare the accuracy of these two estimators that have the same cost in terms of needed data versus the reference MC solution obtained with $20,000$ samples. We observe that, for the pressure, the PDFs obtained from the surrogate model at all $3$ time instances are almost identical to the reference solution. In addition, the bimodal nature of the PDFs of CO$_2$ saturation is successfully captured. The slight mismatch is mainly caused by the error in predicting the location of the discontinuity front. Again, the accuracy of the PDFs at the time instance where no training data were provided (e.g., $150$ days) is visually the same to that at the training time instances (i.e., $100$  and $200$ days). Note that to obtain an almost identical result to the reference solution, the deep network-based MC simulation only uses $1600$ model runs for surrogate construction. On the other hand, with the same $1600$ model runs, the traditional MC obtains the estimated PDFs that show a large approximation error (the blue dot dash lines) especially for the bimodal PDFs of saturation.

\begin{figure}[h!]
\centering
\includegraphics[width=0.9\textwidth]{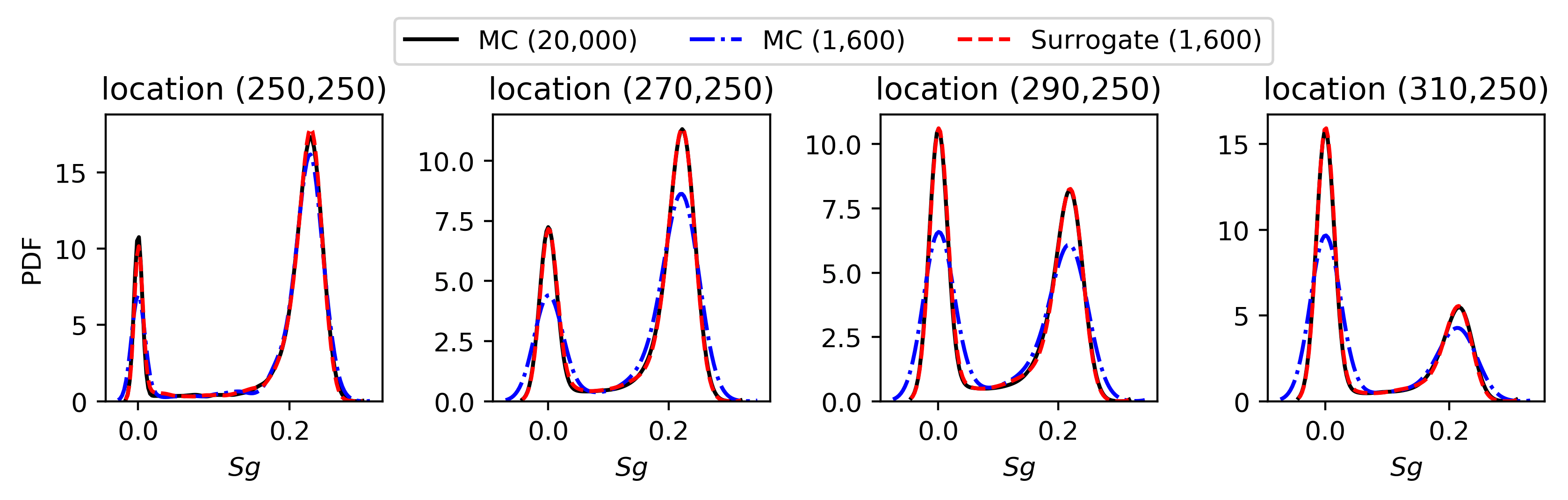}
    \caption{PDFs of CO$_2$ saturation ($Sg$) at four locations at $200$ days. The numbers of model runs needed by MC and the surrogate method are shown in parenthesis.}
\label{PDF_Sg_200d}
\end{figure}

\begin{figure}[h!]
\centering
\includegraphics[width=0.9\textwidth]{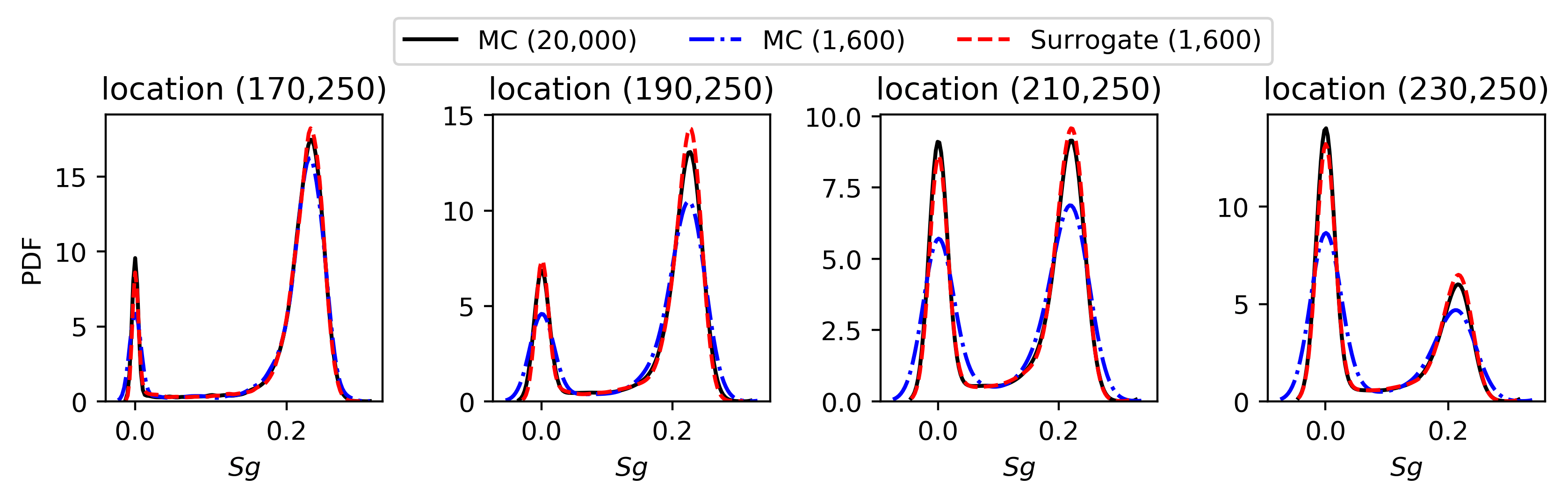}
    \caption{PDFs of CO$_2$ saturation ($Sg$) at four locations at $150$ days. The numbers of model runs needed by MC and the surrogate method are shown in parenthesis. Note that no training data have been provided at $150$ days.}
\label{PDF_Sg_150d}
\end{figure}

The model's performance in capturing the bimodal PDFs of saturation is further illustrated in Figures~\ref{PDF_Sg_200d} and~\ref{PDF_Sg_150d}, which depict the PDFs of CO$_2$ saturation at four locations along the $y$-axis at the time instances $200$ and $150$ days, respectively. It is observed that the model accurately reproduces the bimodal PDFs at both time instances. By treating the time as an additional input and training the network with the outputs at a limited number of time instances, the model can make predictions at arbitrary time instances, enabling the characterization of dynamic systems. The results indicate that the proposed method can efficiently provide accurate solutions of UQ for the dynamic multiphase flow GCS model in the case of high-dimensionality and response discontinuity. 

\section{Conclusions}\label{section:conc}

Surrogate methods are used widely to alleviate the large computational burden associated with uncertainty quantification. However, there are situations that hinder their application to transient multi-phase flow problems due to the high-dimensionality and response discontinuity. In this study, a new deep convolutional neural network approach is proposed for such dynamic problems using an encoder-decoder network architecture that transformed the surrogate modeling problem to an image-to-image regression problem. The encoder is used to extract the underlying features from the high-dimensional input images (fields) which are subsequently used by the decoder to reconstruct the output images (responses). In order to address the need for big data in training deep learning models, a dense fully convolutional neural network (CNN) structure is employed.  It is well-suitable for image processing since it substantially strengthens the information flow through the network. The deep network structure based on CNNs implicitly reduces the input dimensionality through a series of inherent nonlinear projections of the input into high-level coarse feature maps while providing a robust capability to handle complex output responses.

In addition, a training strategy combining a regression loss and a segmentation loss is introduced in order to better approximate the discontinuous saturation field. The segmentation loss is mostly induced by the mismatch in regions around the discontinuous saturation front, thus facilitating a better approximation of the discontinuity. To address the dynamic nature of the problem, time is treated as an extra input to the network that is trained using model outputs at a limited number of time instances.

The performance of the  method developed is demonstrated using a geological carbon storage   process-based multiphase flow model with $2500$ uncertain input parameters. The results indicate that the image-to-image regression strategy is robust in handling problems with high-dimensional input and output fields  with discontinuous response. The method   successfully reproduces the saturation discontinuity and provides accurate approximations of the pressure and saturation fields. Moreover, it is able to characterize the time-dependent multi-output of the dynamic system at arbitrary time instances. The application of the method in addressing uncertainty quantification tasks for  the GCS model showed that the deep network based surrogate model can achieve comparably accurate results to those obtained with traditional Monte Carlo sampling but at a much higher efficiency requiring significantly fewer GCS model runs.

It is worth noting that the ability of the deep network approach to model the dynamic behavior of complex systems  is not only useful for uncertainty quantification problems but also holds promise for experimental design tasks or for the solution of  inverse problems~\citep [e.g.,][]{zhang2015,zhang2016}. 
The model presented can be extended to multiple input fields (e.g. permeability and porosity) without any modifications of the network architecture. Its potential use as a surrogate model for many  complex systems beyond groundwater multi-phase flows remains to be explored.

\acknowledgments
 S.M. acknowledges the China Scholarship Council for financially supporting his study at the Center for Informatics and Computational Science (CICS) at the University of Notre Dame.  N.Z. and Y.Z. acknowledge support from the University of Notre Dame and CICS. The work of S.M., X.S. and J.W. was supported by the National Natural Science Foundation of China (No.  U$1503282$ and $41672229$). The Python codes and data used will be made available at \url{https://github.com/cics-nd/dcedn-gcs} upon publication of this manuscript. The data used can be reproduced using TOUGH2 (\url{http://esd1.lbl.gov/research/projects/tough/}) and the needed scripts are available upon request from the corresponding authors. 


\begin{thebibliography}{60}
	\providecommand{\natexlab}[1]{#1}
	\expandafter\ifx\csname urlstyle\endcsname\relax
	\providecommand{\doi}[1]{doi:\discretionary{}{}{}#1}\else
	\providecommand{\doi}{doi:\discretionary{}{}{}\begingroup
		\urlstyle{rm}\Url}\fi
	
	\bibitem[{\textit{Al-Rfou et~al.}(2016)\textit{Al-Rfou, Alain, Almahairi,
			Angermueller, Bahdanau, Ballas, Bastien, Bayer, Belikov, Belopolsky
			et~al.}}]{al2016theano}
	Al-Rfou, R., G.~Alain, A.~Almahairi, C.~Angermueller, D.~Bahdanau, N.~Ballas,
	F.~Bastien, J.~Bayer, A.~Belikov, A.~Belopolsky, et~al. (2016), Theano: A
	python framework for fast computation of mathematical expressions,
	\textit{arXiv preprint arXiv:1605.02688}, \textit{472}, 473.
	
	\bibitem[{\textit{Asher et~al.}(2015)\textit{Asher, Croke, Jakeman, and
			Peeters}}]{Asher2015}
	Asher, M.~J., B.~F.~W. Croke, A.~J. Jakeman, and L.~J.~M. Peeters (2015), A
	review of surrogate models and their application to groundwater modeling,
	\textit{Water Resources Research}, \textit{51}(8), 5957--5973,
	\doi{10.1002/2015WR016967}.
	
	\bibitem[{\textit{Badrinarayanan et~al.}(2017)\textit{Badrinarayanan, Kendall,
			and Cipolla}}]{badrinarayanan2017}
	Badrinarayanan, V., A.~Kendall, and R.~Cipolla (2017), Segnet: A deep
	convolutional encoder-decoder architecture for image segmentation,
	\textit{IEEE transactions on pattern analysis and machine intelligence},
	\textit{39}(12), 2481--2495.
	
	\bibitem[{\textit{Ballio and Guadagnini}(2004)}]{Ballio2004}
	Ballio, F., and A.~Guadagnini (2004), Convergence assessment of numerical monte
	carlo simulations in groundwater hydrology, \textit{Water Resources
		Research}, \textit{40}(4), n/a--n/a, \doi{10.1029/2003WR002876}, w04603.
	
	\bibitem[{\textit{Benson and Orr}(2008)}]{benson2008}
	Benson, S.~M., and F.~M. Orr (2008), Carbon dioxide capture and storage,
	\textit{MRS Bulletin}, \textit{33}(4), 303--305, \doi{10.1557/mrs2008.63}.
	
	\bibitem[{\textit{Birkholzer et~al.}(2015)\textit{Birkholzer, Oldenburg, and
			Zhou}}]{BIRKHOLZER2015}
	Birkholzer, J.~T., C.~M. Oldenburg, and Q.~Zhou (2015), Co$_2$ migration and
	pressure evolution in deep saline aquifers, \textit{International Journal of
		Greenhouse Gas Control}, \textit{40}, 203 -- 220,
	\doi{https://doi.org/10.1016/j.ijggc.2015.03.022}, special Issue
	commemorating the 10th year anniversary of the publication of the
	Intergovernmental Panel on Climate Change Special Report on CO$_2$ Capture
	and Storage.
	
	\bibitem[{\textit{Celia et~al.}(2015)\textit{Celia, Bachu, Nordbotten, and
			Bandilla}}]{Celia2015}
	Celia, M.~A., S.~Bachu, J.~M. Nordbotten, and K.~W. Bandilla (2015), Status of
	co$_2$ storage in deep saline aquifers with emphasis on modeling approaches
	and practical simulations, \textit{Water Resources Research}, \textit{51}(9),
	6846--6892, \doi{10.1002/2015WR017609}.
	
	\bibitem[{\textit{Chan and Elsheikh}(2017)}]{chan2017}
	Chan, S., and A.~H. Elsheikh (2017), Parametrization and generation of
	geological models with generative adversarial networks, \textit{arXiv
		preprint arXiv:1708.01810}.
	
	\bibitem[{\textit{Chan and Elsheikh}(2018)}]{CHAN2018}
	Chan, S., and A.~H. Elsheikh (2018), A machine learning approach for efficient
	uncertainty quantification using multiscale methods, \textit{Journal of
		Computational Physics}, \textit{354}, 493 -- 511,
	\doi{https://doi.org/10.1016/j.jcp.2017.10.034}.
	
	\bibitem[{\textit{Cihan et~al.}(2013)\textit{Cihan, Birkholzer, and
			Zhou}}]{Cihan2013}
	Cihan, A., J.~T. Birkholzer, and Q.~Zhou (2013), Pressure buildup and brine
	migration during co$_2$ storage in multilayered aquifers,
	\textit{Groundwater}, \textit{51}(2), 252--267,
	\doi{10.1111/j.1745-6584.2012.00972.x}.
	
	\bibitem[{\textit{Corey et~al.}(1954)}]{corey1954}
	Corey, A.~T., et~al. (1954), The interrelation between gas and oil relative
	permeabilities, \textit{Producers monthly}, \textit{19}(1), 38--41.
	
	\bibitem[{\textit{Crevill{\' e}én-Garc{\' i}a et~al.}(2017)\textit{Crevill{\'
				e}én-Garc{\' i}a, Wilkinson, Shah, and Power}}]{CREVILLENGARCIA20171}
	Crevill{\' e}én-Garc{\' i}a, D., R.~Wilkinson, A.~Shah, and H.~Power (2017),
	Gaussian process modelling for uncertainty quantification in
	convectively-enhanced dissolution processes in porous media, \textit{Advances
		in Water Resources}, \textit{99}, 1 -- 14,
	\doi{https://doi.org/10.1016/j.advwatres.2016.11.006}.
	
	\bibitem[{\textit{Dumoulin and Visin}(2016)}]{dumoulin2016}
	Dumoulin, V., and F.~Visin (2016), A guide to convolution arithmetic for deep
	learning, \textit{arXiv preprint arXiv:1603.07285}.
	
	\bibitem[{\textit{Ganapathysubramanian and
			Zabaras}(2007)}]{GANAPATHYSUBRAMANIAN2007652}
	Ganapathysubramanian, B., and N.~Zabaras (2007), Sparse grid collocation
	schemes for stochastic natural convection problems, \textit{Journal of
		Computational Physics}, \textit{225}(1), 652 -- 685,
	\doi{https://doi.org/10.1016/j.jcp.2006.12.014}.
	
	\bibitem[{\textit{Goodfellow et~al.}(2016)\textit{Goodfellow, Bengio, and
			Courville}}]{Goodfellow-et-al-2016}
	Goodfellow, I., Y.~Bengio, and A.~Courville (2016), \textit{Deep Learning}, MIT
	Press.
	
	\bibitem[{\textit{Gu et~al.}(2018)\textit{Gu, Wang, Kuen, Ma, Shahroudy, Shuai,
			Liu, Wang, Wang, Cai, and Chen}}]{GU2018}
	Gu, J., Z.~Wang, J.~Kuen, L.~Ma, A.~Shahroudy, B.~Shuai, T.~Liu, X.~Wang,
	G.~Wang, J.~Cai, and T.~Chen (2018), Recent advances in convolutional neural
	networks, \textit{Pattern Recognition}, \textit{77}, 354 -- 377,
	\doi{https://doi.org/10.1016/j.patcog.2017.10.013}.
	
	\bibitem[{\textit{He et~al.}(2016)\textit{He, Zhang, Ren, and Sun}}]{he2016}
	He, K., X.~Zhang, S.~Ren, and J.~Sun (2016), Deep residual learning for image
	recognition, in \textit{Proceedings of the IEEE conference on computer vision
		and pattern recognition}, pp. 770--778.
	
	\bibitem[{\textit{Hornik et~al.}(1989)\textit{Hornik, Stinchcombe, and
			White}}]{HORNIK1989}
	Hornik, K., M.~Stinchcombe, and H.~White (1989), Multilayer feedforward
	networks are universal approximators, \textit{Neural Networks},
	\textit{2}(5), 359 -- 366,
	\doi{https://doi.org/10.1016/0893-6080(89)90020-8}.
	
	\bibitem[{\textit{Huang et~al.}(2017)\textit{Huang, Liu, Weinberger, and
			van~der Maaten}}]{huang2017}
	Huang, G., Z.~Liu, K.~Q. Weinberger, and L.~van~der Maaten (2017), Densely
	connected convolutional networks, in \textit{Proceedings of the IEEE
		conference on computer vision and pattern recognition}, vol.~1, p.~3.
	
	\bibitem[{\textit{Ioffe and Szegedy}(2015)}]{ioffe2015batch}
	Ioffe, S., and C.~Szegedy (2015), Batch normalization: Accelerating deep
	network training by reducing internal covariate shift, \textit{arXiv preprint
		arXiv:1502.03167}.
	
	\bibitem[{\textit{Kingma and Ba}(2014)}]{Kingma2014}
	Kingma, D.~P., and J.~Ba (2014), Adam: {A} method for stochastic optimization,
	\textit{CoRR}, \textit{abs/1412.6980}.
	
	\bibitem[{\textit{Kitanidis}(2015)}]{Kitanidis2015}
	Kitanidis, P.~K. (2015), Persistent questions of heterogeneity, uncertainty,
	and scale in subsurface flow and transport, \textit{Water Resources
		Research}, \textit{51}(8), 5888--5904, \doi{10.1002/2015WR017639}.
	
	\bibitem[{\textit{Laloy et~al.}(2013)\textit{Laloy, Rogiers, Vrugt, Mallants,
			and Jacques}}]{Laloy2013}
	Laloy, E., B.~Rogiers, J.~A. Vrugt, D.~Mallants, and D.~Jacques (2013),
	Efficient posterior exploration of a high-dimensional groundwater model from
	two-stage markov chain monte carlo simulation and polynomial chaos expansion,
	\textit{Water Resources Research}, \textit{49}(5), 2664--2682,
	\doi{10.1002/wrcr.20226}.
	
	\bibitem[{\textit{Laloy et~al.}(2017)\textit{Laloy, H{\' e}rault, Lee, Jacques,
			and Linde}}]{LALOY2017387}
	Laloy, E., R.~H{\' e}rault, J.~Lee, D.~Jacques, and N.~Linde (2017), Inversion
	using a new low-dimensional representation of complex binary geological media
	based on a deep neural network, \textit{Advances in Water Resources},
	\textit{110}, 387 -- 405,
	\doi{https://doi.org/10.1016/j.advwatres.2017.09.029}.
	
	\bibitem[{\textit{Laloy et~al.}(2018)\textit{Laloy, H{\' e}rault, Jacques, and
			Linde}}]{LALOY2018}
	Laloy, E., R.~H{\' e}rault, D.~Jacques, and N.~Linde (2018), Training-image
	based geostatistical inversion using a spatial generative adversarial neural
	network, \textit{Water Resources Research}, pp. n/a--n/a,
	\doi{10.1002/2017WR022148}.
	
	\bibitem[{\textit{Li and Zhang}(2007)}]{LiZhang2007}
	Li, H., and D.~Zhang (2007), Probabilistic collocation method for flow in
	porous media: Comparisons with other stochastic methods, \textit{Water
		Resources Research}, \textit{43}(9), n/a--n/a, \doi{10.1029/2006WR005673},
	w09409.
	
	\bibitem[{\textit{Li et~al.}(2009)\textit{Li, Lu, and Zhang}}]{LI2009}
	Li, W., Z.~Lu, and D.~Zhang (2009), Stochastic analysis of unsaturated flow
	with probabilistic collocation method, \textit{Water Resources Research},
	\textit{45}(8), n/a--n/a, \doi{10.1029/2008WR007530}, w08425.
	
	\bibitem[{\textit{Li et~al.}(2017)\textit{Li, Kokkinaki, Darve, and
			Kitanidis}}]{LiYJ2017}
	Li, Y.~J., A.~Kokkinaki, E.~F. Darve, and P.~K. Kitanidis (2017),
	Smoothing-based compressed state kalman filter for joint state-parameter
	estimation: Applications in reservoir characterization and co$_2$ storage
	monitoring, \textit{Water Resources Research}, \textit{53}(8), 7190--7207,
	\doi{10.1002/2016WR020168}.
	
	\bibitem[{\textit{Liao and Zhang}(2013)}]{LiaoZhang2013}
	Liao, Q., and D.~Zhang (2013), Probabilistic collocation method for strongly
	nonlinear problems: 1. transform by location, \textit{Water Resources
		Research}, \textit{49}(12), 7911--7928, \doi{10.1002/2013WR014055}.
	
	\bibitem[{\textit{Liao and Zhang}(2014)}]{LiaoZhang2014}
	Liao, Q., and D.~Zhang (2014), Probabilistic collocation method for strongly
	nonlinear problems: 2. transform by displacement, \textit{Water Resources
		Research}, \textit{50}(11), 8736--8759, \doi{10.1002/2014WR016238}.
	
	\bibitem[{\textit{Liao and Zhang}(2016)}]{LiaoZhang2016}
	Liao, Q., and D.~Zhang (2016), Probabilistic collocation method for strongly
	nonlinear problems: 3. transform by time, \textit{Water Resources Research},
	\textit{52}(3), 2366--2375, \doi{10.1002/2015WR017724}.
	
	\bibitem[{\textit{Liao et~al.}(2017)\textit{Liao, Zhang, and
			Tchelepi}}]{LIAO2017828}
	Liao, Q., D.~Zhang, and H.~Tchelepi (2017), A two-stage adaptive stochastic
	collocation method on nested sparse grids for multiphase flow in randomly
	heterogeneous porous media, \textit{Journal of Computational Physics},
	\textit{330}, 828 -- 845, \doi{https://doi.org/10.1016/j.jcp.2016.10.061}.
	
	\bibitem[{\textit{Lin and Tartakovsky}(2009)}]{LIN2009712}
	Lin, G., and A.~Tartakovsky (2009), An efficient, high-order probabilistic
	collocation method on sparse grids for three-dimensional flow and solute
	transport in randomly heterogeneous porous media, \textit{Advances in Water
		Resources}, \textit{32}(5), 712 -- 722,
	\doi{https://doi.org/10.1016/j.advwatres.2008.09.003}.
	
	\bibitem[{\textit{Long et~al.}(2015)\textit{Long, Shelhamer, and
			Darrell}}]{long2015}
	Long, J., E.~Shelhamer, and T.~Darrell (2015), Fully convolutional networks for
	semantic segmentation, in \textit{Proceedings of the IEEE conference on
		computer vision and pattern recognition}, pp. 3431--3440.
	
	\bibitem[{\textit{Lu et~al.}(2016)\textit{Lu, Zhang, Webster, and
			Barbier}}]{Luetal2016}
	Lu, D., G.~Zhang, C.~Webster, and C.~Barbier (2016), An improved multilevel
	monte carlo method for estimating probability distribution functions in
	stochastic oil reservoir simulations, \textit{Water Resources Research},
	\textit{52}(12), 9642--9660, \doi{10.1002/2016WR019475}.
	
	\bibitem[{\textit{Lu et~al.}(2018)\textit{Lu, Ricciuto, and Evans}}]{lu2018}
	Lu, D., D.~Ricciuto, and K.~Evans (2018), An efficient bayesian data-worth
	analysis using a multilevel monte carlo method, \textit{Advances in Water
		Resources}, \textit{113}, 223--235,
	\doi{https://doi.org/10.1016/j.advwatres.2018.01.024}.
	
	\bibitem[{\textit{Ma and Zabaras}(2009)}]{MA20093084}
	Ma, X., and N.~Zabaras (2009), An adaptive hierarchical sparse grid collocation
	algorithm for the solution of stochastic differential equations,
	\textit{Journal of Computational Physics}, \textit{228}(8), 3084 -- 3113,
	\doi{https://doi.org/10.1016/j.jcp.2009.01.006}.
	
	\bibitem[{\textit{Ma and Zabaras}(2010)}]{MA20103884}
	Ma, X., and N.~Zabaras (2010), An adaptive high-dimensional stochastic model
	representation technique for the solution of stochastic partial differential
	equations, \textit{Journal of Computational Physics}, \textit{229}(10), 3884
	-- 3915, \doi{https://doi.org/10.1016/j.jcp.2010.01.033}.
	
	\bibitem[{\textit{Meng and Li}(2017)}]{MENG201713}
	Meng, J., and H.~Li (2017), An efficient stochastic approach for flow in porous
	media via sparse polynomial chaos expansion constructed by feature selection,
	\textit{Advances in Water Resources}, \textit{105}, 13 -- 28,
	\doi{https://doi.org/10.1016/j.advwatres.2017.04.019}.
	
	\bibitem[{\textit{Mo et~al.}(2017)\textit{Mo, Lu, Shi, Zhang, Ye, Wu, and
			Wu}}]{MO2017}
	Mo, S., D.~Lu, X.~Shi, G.~Zhang, M.~Ye, J.~Wu, and J.~Wu (2017), A taylor
	expansion-based adaptive design strategy for global surrogate modeling with
	applications in groundwater modeling, \textit{Water Resources Research},
	\textit{53}(12), 10,802--10,823, \doi{10.1002/2017WR021622}.
	
	\bibitem[{\textit{Mualem}(1976)}]{Mualem1976}
	Mualem, Y. (1976), A new model for predicting the hydraulic conductivity of
	unsaturated porous media, \textit{Water Resources Research}, \textit{12}(3),
	513--522, \doi{10.1029/WR012i003p00513}.
	
	\bibitem[{\textit{M{\"u}ller et~al.}(2011)\textit{M{\"u}ller, Jenny, and
			Meyer}}]{MULLER20111527}
	M{\"u}ller, F., P.~Jenny, and D.~W. Meyer (2011), Probabilistic collocation and
	lagrangian sampling for advective tracer transport in randomly heterogeneous
	porous media, \textit{Advances in Water Resources}, \textit{34}(12), 1527 --
	1538, \doi{https://doi.org/10.1016/j.advwatres.2011.09.005}.
	
	\bibitem[{\textit{Pruess}(2005)}]{pruess2005}
	Pruess, K. (2005), \textit{ECO2N: A TOUGH2 fluid property module for mixtures
		of water, NaCl, and CO$_2$}, Lawrence Berkeley National Laboratory Berkeley,
	Berkeley, CA, USA.
	
	\bibitem[{\textit{Pruess et~al.}(1999)\textit{Pruess, Oldenburg, and
			Moridis}}]{pruess1999}
	Pruess, K., C.~Oldenburg, and G.~Moridis (1999), Tough2 user's guide version 2,
	\textit{Tech. rep.}, Lawrence Berkeley National Laboratory, Berkeley, CA,
	USA.
	
	\bibitem[{\textit{Rasmussen and Williams}(2006)}]{rasmussen2006}
	Rasmussen, C.~E., and C.~K.~I. Williams (2006), \textit{Gaussian Processes for
		Machine Learning}, MIT Press.
	
	\bibitem[{\textit{Razavi et~al.}(2012)\textit{Razavi, Tolson, and
			Burn}}]{Razavi2012}
	Razavi, S., B.~A. Tolson, and D.~H. Burn (2012), Review of surrogate modeling
	in water resources, \textit{Water Resources Research}, \textit{48}(7),
	n/a--n/a, \doi{10.1029/2011WR011527}, w07401.
	
	\bibitem[{\textit{Tian et~al.}(2017)\textit{Tian, Wilkinson, Yang, Power,
			Fagerlund, and Niemi}}]{TIAN2017113}
	Tian, L., R.~Wilkinson, Z.~Yang, H.~Power, F.~Fagerlund, and A.~Niemi (2017),
	Gaussian process emulators for quantifying uncertainty in co$_2$ spreading
	predictions in heterogeneous media, \textit{Computers \& Geosciences},
	\textit{105}, 113 -- 119, \doi{https://doi.org/10.1016/j.cageo.2017.04.006}.
	
	\bibitem[{\textit{Tripathy and Bilionis}(2018)}]{tripathy2018}
	Tripathy, R., and I.~Bilionis (2018), Deep uq: Learning deep neural network
	surrogate models for high dimensional uncertainty quantification,
	\textit{arXiv preprint arXiv:1802.00850}.
	
	\bibitem[{\textit{van Genuchten}(1980)}]{van1980}
	van Genuchten, M.~T. (1980), A closed-form equation for predicting the
	hydraulic conductivity of unsaturated soils 1, \textit{Soil science society
		of America journal}, \textit{44}(5), 892--898.
	
	\bibitem[{\textit{Xiu and Hesthaven}(2005)}]{Xiu2005}
	Xiu, D., and J.~S. Hesthaven (2005), High-order collocation methods for
	differential equations with random inputs, \textit{SIAM Journal on Scientific
		Computing}, \textit{27}(3), 1118--1139, \doi{10.1137/040615201}.
	
	\bibitem[{\textit{Xiu and Karniadakis}(2002)}]{xiu2002}
	Xiu, D., and G.~E. Karniadakis (2002), The wiener--askey polynomial chaos for
	stochastic differential equations, \textit{SIAM Journal on Scientific
		Computing}, \textit{24}(2), 619--644, \doi{10.1137/S1064827501387826}.
	
	\bibitem[{\textit{Zhang and Lu}(2004)}]{ZHANG2004773}
	Zhang, D., and Z.~Lu (2004), An efficient, high-order perturbation approach for
	flow in random porous media via karhunen-lo{\` e}ve and polynomial
	expansions, \textit{Journal of Computational Physics}, \textit{194}(2), 773
	-- 794, \doi{https://doi.org/10.1016/j.jcp.2003.09.015}.
	
	\bibitem[{\textit{Zhang et~al.}(2013)\textit{Zhang, Lu, Ye, Gunzburger, and
			Webster}}]{ZHANG2013}
	Zhang, G., D.~Lu, M.~Ye, M.~Gunzburger, and C.~Webster (2013), An adaptive
	sparse-grid high-order stochastic collocation method for bayesian inference
	in groundwater reactive transport modeling, \textit{Water Resources
		Research}, \textit{49}(10), 6871--6892, \doi{10.1002/wrcr.20467}.
	
	\bibitem[{\textit{Zhang et~al.}(2015)\textit{Zhang, Zeng, Chen, Chen, and
			Wu}}]{zhang2015}
	Zhang, J., L.~Zeng, C.~Chen, D.~Chen, and L.~Wu (2015), Efficient bayesian
	experimental design for contaminant source identification, \textit{Water
		Resources Research}, \textit{51}(1), 576--598, \doi{10.1002/2014WR015740}.
	
	\bibitem[{\textit{Zhang et~al.}(2016)\textit{Zhang, Li, Zeng, and
			Wu}}]{zhang2016}
	Zhang, J., W.~Li, L.~Zeng, and L.~Wu (2016), An adaptive gaussian process-based
	method for efficient bayesian experimental design in groundwater contaminant
	source identification problems, \textit{Water Resources Research},
	\textit{52}(8), 5971--5984, \doi{10.1002/2016WR018598}.
	
	\bibitem[{\textit{Zhu and Zabaras}(2018)}]{zhu2018}
	Zhu, Y., and N.~Zabaras (2018), Bayesian deep convolutional encoder-decoder
	networks for surrogate modeling and uncertainty quantification, \textit{Journal of Computational Physics}, \textit{366}, 415--447, \doi{https://doi.org/10.1016/j.jcp.2018.04.018}.
	
\end{thebibliography}

\listofchanges

\end{document}